\documentclass[conference]{IEEEtran}
\IEEEoverridecommandlockouts
\usepackage{cite}
\usepackage{amsmath,amssymb,amsfonts}
\usepackage{algorithmic}
\usepackage{graphicx}
\usepackage{textcomp}
\usepackage{xcolor}
\usepackage{float}
\usepackage{booktabs}
\usepackage{hyperref}
\usepackage{multirow}

\def\BibTeX{{\rm B\kern-.05em{\sc i\kern-.025em b}\kern-.08em
    T\kern-.1667em\lower.7ex\hbox{E}\kern-.125emX}}
\begin{document}

\title{Spatial-Assistant Encoder-Decoder Network for Real Time Semantic Segmentation}

 \author{
 \IEEEauthorblockN{Yalun Wang ,Shidong Chen ,Huicong Bian ,Weixiao Li ,Qin Lu}
}
\maketitle

\begin{abstract}

Semantic segmentation is an essential technology for self-driving cars to comprehend their surroundings. Currently, real-time semantic segmentation networks commonly employ either encoder-decoder architecture or two-pathway architecture. Generally speaking, encoder-decoder models tend to be quicker, whereas two-pathway models exhibit higher accuracy. To leverage both strengths, we present the Spatial-Assistant Encoder-Decoder Network (SANet) to fuse the two architectures. In the overall architecture, we uphold the encoder-decoder design while maintaining the feature maps in the middle section of the encoder and utilizing atrous convolution branches for same-resolution feature extraction. Toward the end of the encoder, we integrate the asymmetric pooling pyramid pooling module (APPPM) to optimize the semantic extraction of the feature maps. This module incorporates asymmetric pooling layers that extract features at multiple resolutions. In the decoder, we present a hybrid attention module, SAD, that integrates horizontal and vertical attention to facilitate the combination of various branches. To ascertain the effectiveness of our approach, our SANet model achieved competitive results on the real-time CamVid and cityscape datasets. By employing a single 2080Ti GPU, SANet achieved a 78.4 $\%$  mIOU at 65.1 FPS on the Cityscape test dataset and 78.8 $\%$  mIOU at 147 FPS on the CamVid test dataset. The training code and model for SANet are available at \href{https://github.com/CuZaoo/SANet-main}{https://github.com/CuZaoo/SANet-main}

\end{abstract}

\section{\textbf{Introduction}}
Real-time semantic segmentation is a crucial research topic in computer vision. It aims to classify each pixel in an image at real-time speed (FPS $\ge $ 30), thus enabling pixel-level semantic segmentation. This task hinges on assigning each pixel in an image to a specific semantic category, such as people, vehicles, roads, trees, etc., in order to provide fine-grained and high-level semantic information about the image. Real-time semantic segmentation has numerous applications in various domains such as autonomous driving\cite{yang2022research}, drone applications\cite{shahmoradi2020comprehensive}, smart healthcare\cite{shen2020smart}, and remote sensing image analysis\cite{diakogiannis2020resunet}. Utilizing real-time semantic segmentation, applications in these domains can understand environments and scenarios more effectively, allowing for more informed and accurate decisions and actions.

Recently, the emergence of deep learning has led to significant advancements in real-time semantic segmentation. The inception of Full Convolutional Neural Networks (FCN)\cite{long2015fully} laid the foundation for achieving real-time semantic segmentation, which achieves end-to-end pixel-level classification via appropriate network design, effectively decreasing the computation and parameter count. Since then, numerous

\begin{figure}[H]
\includegraphics[width=8.5cm,height=8.5cm]{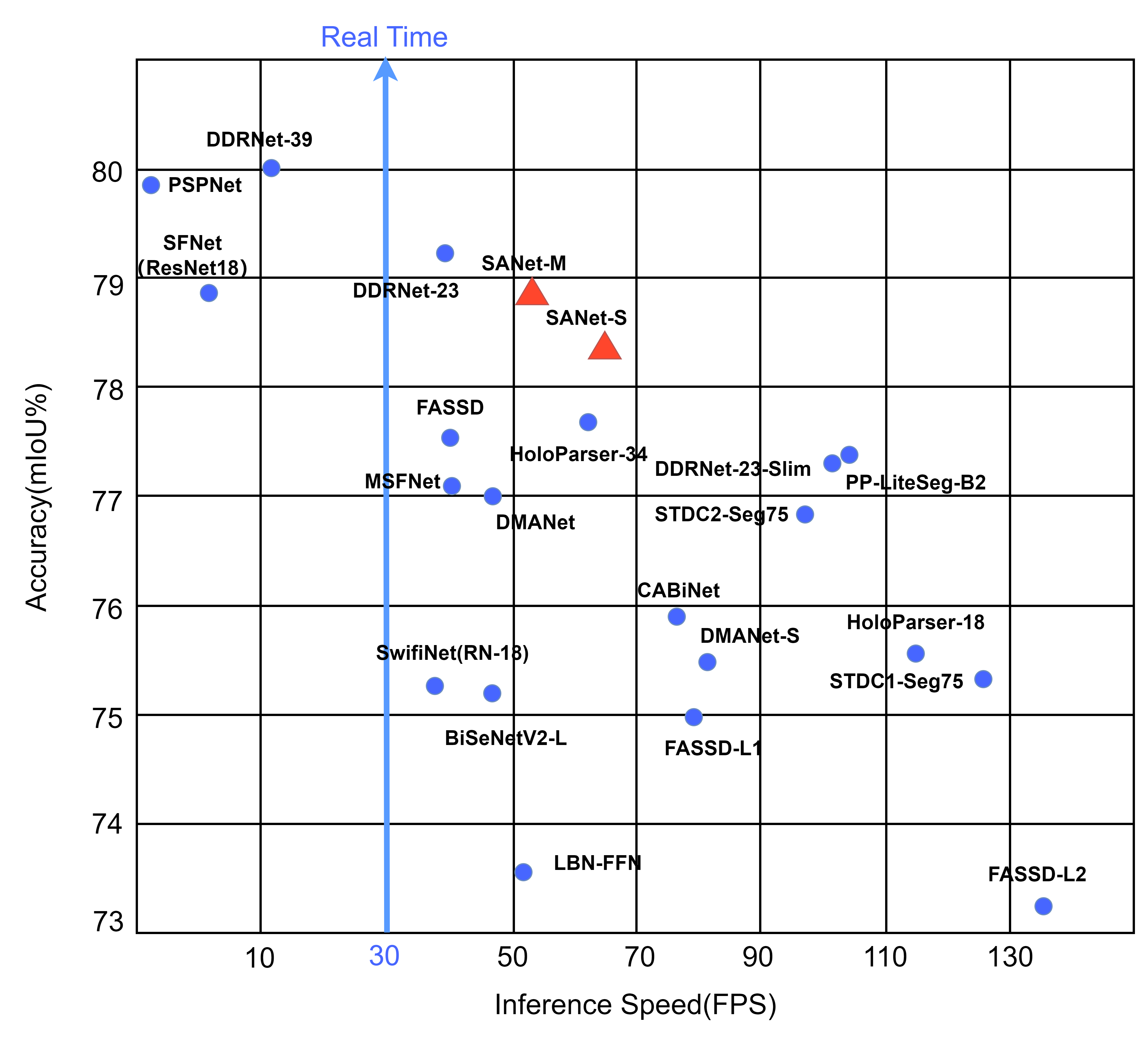}
\caption{Tradeoffs between reasoning speed and accuracy (reporting) of real-time models on the test dataset of Cityscapes. The red triangles represent SANet, and the blue circles represent the other models.}
\label{allnet}
\end{figure}

\begin{table*}[htp]
\caption{\textbf{The structure of SANet's Encoder}. "Output" denotes the dimension of the output feature mapping, and "Repeat" is the number of times the BasicBlock is repeated in a single  Layer.}
\begin{center}
\renewcommand\arraystretch{2}
\label{table1}
\begin{tabular}{|l|c|c|c|c|c|c|c|c|}
\hline
Stage       & \textbf{Stom} & \textbf{Layer1} & \textbf{Layer2} & \textbf{Layer3} & \textbf{Layer4} & \textbf{Layer5} & \textbf{Layer6} & \textbf{APPPM} \\ \hline
Output      & 32$\times$512$\times$256    & 32$\times$512$\times$256      & 64$\times$256$\times$128      & 128$\times$256$\times$128     & 128$\times$128$\times$64      & 256$\times$64$\times$32       & 512$\times$32$\times$16       & 128$\times$32$\times$16      \\ \hline
Repeat(S-M) & 1-1         & 2-3           & 2-3           & 2-3          & \textbf{2-9} & 2-3           & 1-1           &                \\ \hline


\end{tabular}
\end{center}
\end{table*}

\noindent enhanced network structures and algorithms have arisen, including U-Net\cite{ronneberger2015u}, SegNet\cite{badrinarayanan2017segnet}, PSPNet\cite{zhao2017pyramid}, ENet\cite{paszke2016enet}, etc. These have consistently enhanced the performance of real-time semantic segmentation.

However, real-time semantic segmentation continues to encounter several challenges, such as enhancing segmentation precision, expediting processing velocity, and accomplishing real-time segmentation within resource-constrained settings. Hence, it is imperative for real-time semantic segmentation to attain faster processing velocity while maintaining elevated segmentation precision to cater to the real-time applications’ stipulations, which demand further research and
innovation.

To enhance network segmentation performance, researchers have employed multiple strategies. These include designing intricate network structural blocks, extending network branches, and maximizing image feature reuse. Complex network building blocks typically refine the 3$\times$3 convolution into more intricate branches. By extending these branches, the network becomes more parallelized. To increase the multiplexing of image features, a large number of 1$\times$1 convolutions are needed to adjust the number of channels. Although these methods enhance the image segmentation performance of the network, they concurrently augment the network's complexity, thereby potentially hindering the inference speed of the network.

To ameliorate the inference speed of the network, lightweight networks are designed as backbones for extracting image features. However, lighter networks typically possess fewer feature channels and shallower network depth. This may cause the network-extracted image features to be inadequate in enhancing the image segmentation accuracy. As a result, researchers must attain a balance between segmentation performance and processing speed that fulfills the requirements of various application scenarios.

In this paper, we propose a SANet real-time semantic segmentation network, which uses a combination of a primary branch and a subbranches in the encoding phase. The primary branches comprise multiple common convolutional layers that have a stride of 2. These are primarily used to downsample the input image adequately enough to extract the image's semantic information. The sub-branches feature fewer atrous convolution layers and aim to support the primary branches in maintaining the image's spatial information. The incorporation of a pyramid pooling module aids in gathering contextual insights from various feature layers, enhancing the network's capability of gaining global insights. To align with prior research, we introduce a pyramid pooling module (PPM) - the asymmetric pooling pyramid pooling module (APPPM) - at the conclusion of the primary branches.

The primary objective of the decoder is to upsample the high-level characteristics extracted by the encoder to a set resolution. Generally, the crux of the decoder's design is to efficiently recover the compressed high-level semantic features without incurring any loss. To assimilate the semantic and spatial information obtained at the encoder stage, the decoder in SANet amalgamates the received semantic and spatial features.
Based on extensive experiments on Cityscapes\cite{cordts2016cityscapes} and CamVid\cite{brostow2009semantic}, two widely accepted benchmarks, SANet achieves a satisfactory balance between segmentation accuracy and inference speed. Our approach outperforms other real-time semantic segmentation models as we achieve up-to-date accuracy scores on both the Cityscapes and CamVid datasets. Comparative results are illustrated in Figure~\ref{allnet}. With standard test enhancements, SANet can rival state-of-the-art models while utilizing fewer computational resources. After conducting extensive experiments on the CityScapes and CamVid datasets, SANet achieved a mIoU of 78.4$\%$  and FPS of 65.1 on the CityScapes dataset and a mIoU of 78.8$\%$  and FPS of 147 on the CamVid dataset.

The main contributions are summarized as follows: 

\begin{itemize}
\item We introduce a new pyramid pooling module, the APPPM, comprising an asymmetrical pooling layer and a conventional convolutional layer. The asymmetrical pooling layer implements a distinct pooling operation to capture various scale and orientation features of the image. Consequently, this leads to a more even reduction in image resolution, making it easier to extract image features at varying resolutions.
\item We present an effective and straightforward attention decoder called SAD. It enhances the feature representation through hybrid attention, which consists of horizontal and vertical attention, and increases the feature fusion effect using several convolutions.
\item Based on the aforementioned modules, we propose a real-time semantic segmentation model called SANet. Extensive experimentation has demonstrated that SANet strikes a commendable equilibrium between accuracy and speed on both Cityscapes and CamVid. In particular, our SANet obtains a mIoU of 78.4 $\%$ on the Cityscapes test dataset while running at 65.1 FPS on an NVIDIA GTX 2080Ti.
\end{itemize}

\section{\textbf{RELATED WORK}}

\begin{figure*}[h]
\centerline{
 \begin{minipage}{0.3\linewidth}
 	
 	\centerline{\includegraphics[width=\textwidth]{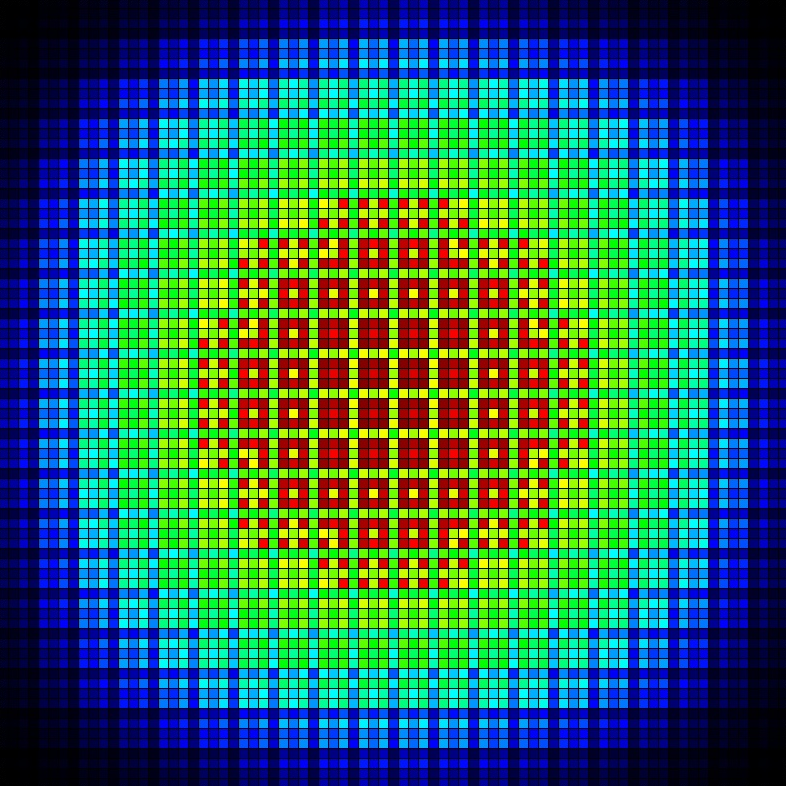}}
 \end{minipage}
 \begin{minipage}{0.3\linewidth}
 	
 	\centerline{\includegraphics[width=\textwidth]{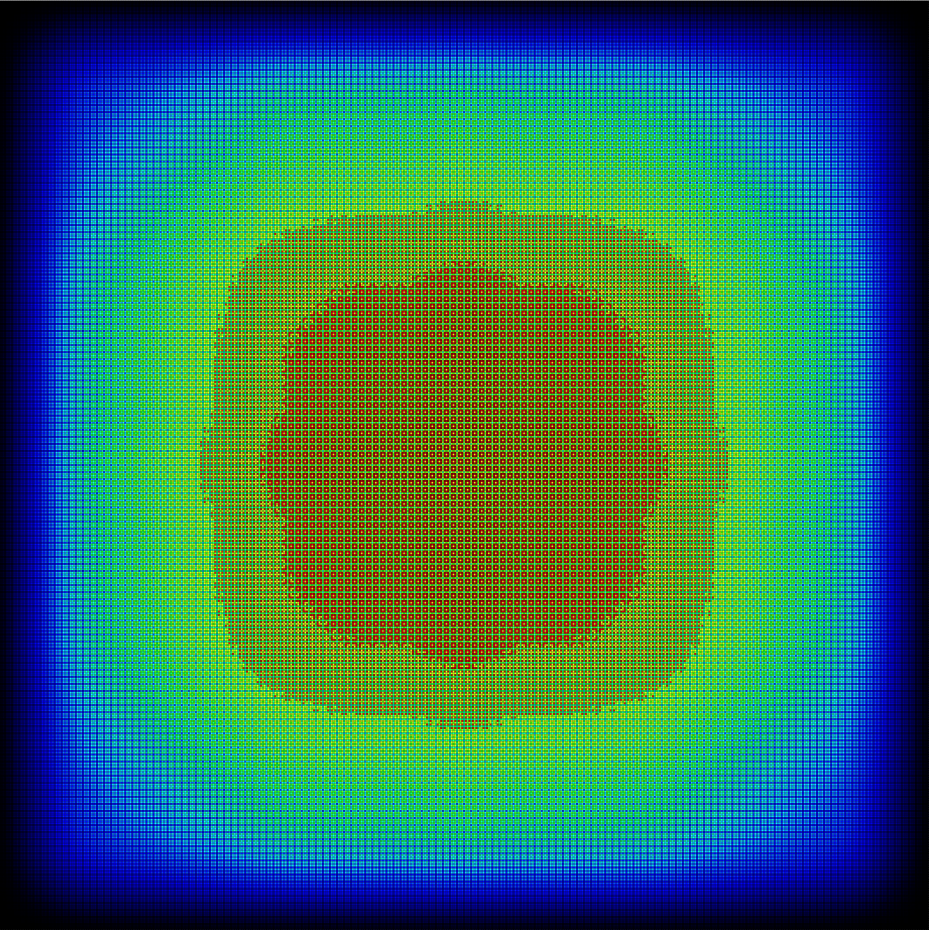}}
 \end{minipage}
 \begin{minipage}{0.3\linewidth}
	
	\centerline{\includegraphics[width=\textwidth]{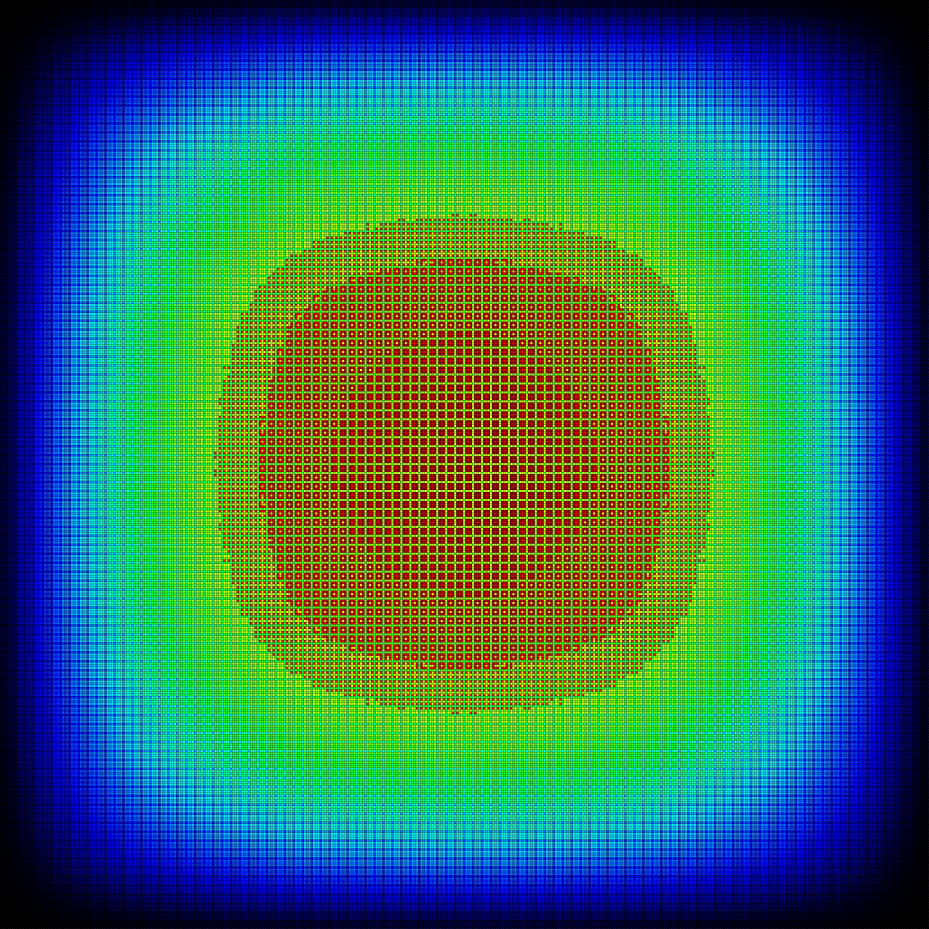}}
\end{minipage}
}

\caption{SANet different periods of receptive field. From left to right: L3, DP2 and L6.}
\label{6}
\end{figure*}

\subsection{\textbf{Semantic Segmentation}}

As deep learning technology continues to advance, semantic segmentation has seen significant progress. Notably among them is FCN\cite{long2015fully}, the foremost model to apply fully convolutional networks to semantic segmentation. Since then, several models have emerged to enhance the performance of semantic segmentation tasks. SegNet\cite{badrinarayanan2017segnet} introduces encoder-decoder architectures that utilize a decoder to recuperate the resolution of an image. To enhance segmentation accuracy and incorporate contextual information, PSPNet\cite{zhao2017pyramid} incorporates a pyramid pooling module(PPM). This module employs pooling layers of varying sizes to expand the receptive field, which facilitates contextual information extraction. The DeepLab\cite{chen2017deeplab}\cite{chen2018encoder} range of models presents various techniques, with a particular focus on the application of atrous convolutions. They also devise a spatial pyramid pooling module (ASPP) to further extract features from the image. HRNet\cite{sun2019high} comprises numerous parallel sub-networks with varying resolutions to address the issue of spatial information loss that arises when reconstructing high-resolution feature maps using low-resolution counterparts. Furthermore, Chen et al\cite{chen2022vision}. improve the processing efficiency of dense prediction tasks by introducing an adapter layer to the current visual Transformer model.

\subsection{\textbf{Real-time Semantic Segmentation}}
Real-time semantic segmentation is a widely researched task that aims to strike a balance between segmentation accuracy and prediction speed. The ultimate goal of real-time semantic segmentation is to achieve high segmentation accuracy in a short amount of time. Currently, research into real-time semantic segmentation is split into two architectures: encoder-decoder architecture and two-pathway architecture.

\paragraph {\textbf{Encoder-decoder Architecture}}Encoder-decoder architectures are widely used in real-time semantic segmentation. The real-time semantic segmentation model, ENet\cite{paszke2016enet}, used an encoder-decoder architecture which, despite its relatively simple structure significantly increased the speed of prediction, but resulted in reduced segmentation accuracy. EDANet\cite{lo2019efficient} introduces a block structure featuring asymmetric and atrous convolution, and enhances the network's prediction accuracy through the employment of dense concatenation. ESNet\cite{wang2019esnet} proposes a novel residual module that utilizes a "transform-split-merge" approach to improve the module's expressive ability. SFNet\cite{li2020semantic}, on the other hand, commences with feature alignment and introduces a Flow Alignment module with a "transform-split-merge" strategy. SFNet initiates with feature alignment by introducing a Flow Alignment Module (FAM) to align the feature maps of adjacent levels for more effective feature fusion. A satisfactory compromise between pace and precision is attained. FANet \cite{tomar2022fanet} achieves a favorable balance between velocity and precision by implementing extra downsampling and fast attention modules in the network. STDC\cite{fan2021rethinking} formulates convolutional units that are better suited for semantic segmentation missions and introduces an auxiliary loss function to enable the network to concentrate on grasping spatial data. DMANet\cite{weng2022deep} develops a decoder for multi-branch aggregation alongside a lattice enhanced residual block to improve the network's feature representation. Atrous convolution assists in enlarging the sensory field and preserving additional spatial information. RegSeg\cite{gao2021rethink} reconceptualises the use of atrous convolution by developing a convolutional block that incorporates atrous convolutions, each with varying dilation rates, and replicating this block structure several times in the principal network for image segmentation. FASSD\cite{rosas2021fassd} uses the convolutional block of HarDNet as Backone, obtains a new module DFPA by replacing the normal convolution in ASPP with an asymmetric convolution and designs a new decoder MDA. This approach enhances model performance by reducing computational complexity, and bridging the accuracy gap between real-time and non-real-time networks.

\begin{figure*}[htp]
\centerline{
\includegraphics[scale=0.12]{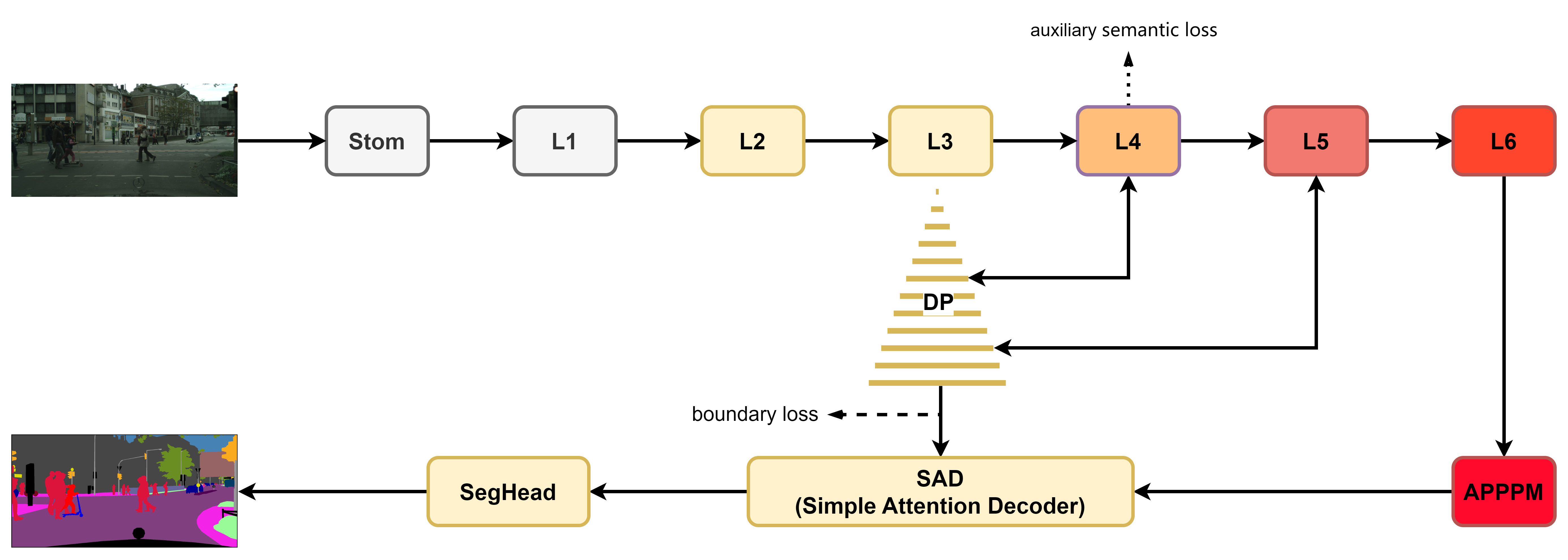}}
\caption{\textbf{Overview of SANet's Basic Structure}: The resolution of different layers [1/4, 1/8, 1/16, 1/32, 1/64] corresponds to the color [white, yellow, orange, light red, dark red]. The heavier the color, the lower the resolution. Following APPPM's extraction of deeper semantic information, SAD reduces the encoder's image information. The additional loss, utilized only during training, comprises auxiliary semantic loss and boundary loss. In this scenario, the auxiliary semantic loss function takes the output of the L4 layer as its input, and the boundary loss takes the output of the DP module as its input.}
\label{net}
\end{figure*}

\paragraph {\textbf{Two-pathway Architecture}}Preserving spatial information in real-time semantic segmentation is a crucial research objective, particularly for smaller objects. Two-pathway architecture is a suggested resolution for this issue. ICNet\cite{zhao2018icnet} receives images with varying resolutions as inputs, directs them into several network branches, and establishes connections through cascading feature fusion units to achieve top-notch segmentation. BiseNet\cite{yu2018bisenet} separates image data into spatial and contextual information and employs less deep spatial paths to obtain the image's spatial information and a deeper semantic path to extract the image's semantic information. These various branches are ultimately integrated into the feature fusion module to achieve information fusion. In the initial stage of the network, DDRNet\cite{pan2022deep} suggests sharing the spatial and contextual paths, followed by splitting the spatial and contextual paths when reducing the image feature map to 1/8 resolution. This technique notably enhances the network's prediction speed. In addition, DDRNet enhances the segmentation accuracy of PPM proposed by PSPNet. DMRNet\cite{wang2023deep}, like DDRNet, utilizes the information of different resolutions but further improves the segmentation accuracy by splitting the spatial and contextual paths at 1/8, 1/16, and 1/32 resolutions. The three branches of PIDNet\cite{xu2023pidnet}, namely P, I, and D, are modeled after PID controllers to address the "overshoot" issue prevalent in two-branch networks. The inclusion of the D (derivative) branch allows for better integration of the P (proportional) and I (integral) branches, thus enhancing the network's performance and stability.

\subsection{\textbf{Semantic Extraction Module}}In semantic segmentation, contextual information can be demonstrated by the extent of the receptive field. Various models aim to enlarge the model's receptive field by implementing a semantic extraction module after downsampling. This is done in hopes of optimizing the use of contextual information. A semantic extraction module comprises multiple parallel convolutional blocks of diverse sizes or pooling layers.

PSPNet\cite{zhao2017pyramid} proposed the Pyramid Pooling Module (PPM) as a multi-scale feature extraction module. The PPM effectively captures semantic information from images at different scales through pyramid pooling operations, thus enhancing semantic segmentation performance. DeepLab V2\cite{chen2017deeplab} proposed the ASPP, which uses atrous convolution for multi-scale sensory field fusion without a pooling operation. The ASPP typically features several convoluted branches running in parallel, each of which adopts a distinct atrous rate to capture information at different scales. The DAPF, which was proposed by FASSD\cite{rosas2021fassd}, utilizes a decomposition technique to break down the 3$\times$3 atrous convolution of the ASPP standard. Splitting the 3$\times$3 atrous convolution into two consecutive asymmetric atrous convolutions reduces computational complexity while preserving the capacity to capture multi-scale contextual information. To enhance detection speed, PPLite's\cite{peng2022pp} SPPM approach reduces the number of channels in the PPM output, removes shortcut connections, and replaces connection operations with addition. Using only one 3$\times$3 or 1$\times$1 convolution in PPM to combine multi-scale context information may not be the most effective approach. To address this, DAPPM, proposed by DDRNet\cite{pan2022deep}, improves the information fusion by adding a 3$\times$3 convolution after the pooling layer. DMRNet\cite{wang2023deep} enhances the module's speed by using asymmetric or atrous convolution instead of 3$\times$3 convolution in DAPPM. PIDNet\cite{xu2023pidnet} parallelizes DAPPM for further enhancement.

Unlike the above feature extraction modules, SANet's feature extraction module APPPM modifies the pooling layer. By modifying the size of the pooling layer, APPPM can extract image features from more resolutions.

\section{METHOD}

In this section, we will first outline the general structure of SANet, which is designed for real-time semantic segmentation. Subsequently, we will delve into the specifics of APPPM and SAD in SANet.

\subsection{\textbf{Design of the Network Structure}}
In terms of the network's general architectural design, SANet utilizes the encoder-decoder approach. The overall structure of the network is shown in Figure~\ref{net}. When it comes

\begin{figure}[H]
\centerline{
\includegraphics[scale=0.11]{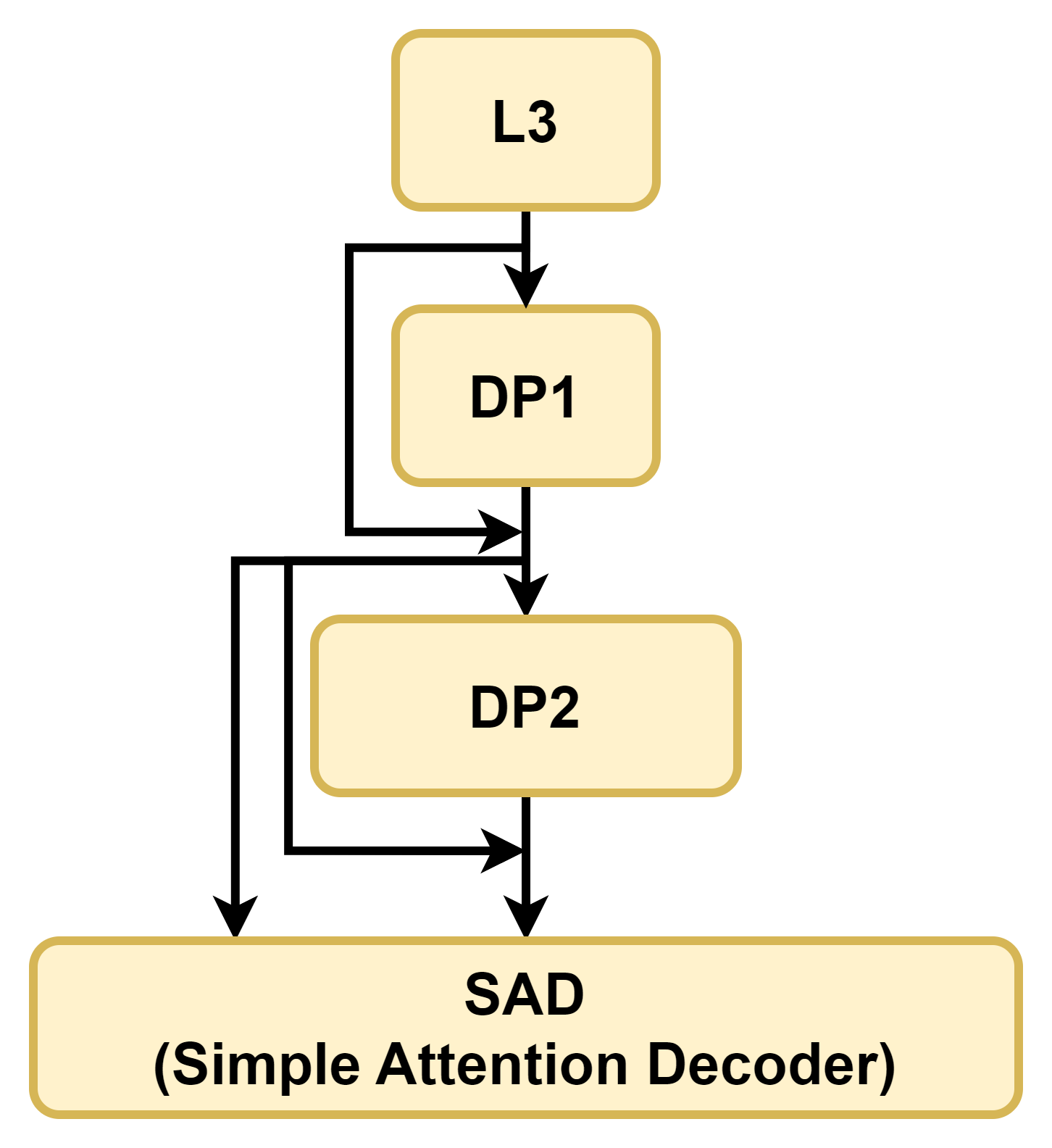}}
\caption{The structure of DP in SANet. In DP, the atrous convolution of DP2 employs a higher atrous rate compared to DP1.}
\label{dp}
\end{figure}

\noindent the encoder, a well-structured convolutional block can significantly enhance the encoder's feature extraction capability. The BasicBlock, employed by ResNet\cite{he2016deep}, is a widely-used choice as it allows for the smooth flow of information throughout the network while preventing the vanishing gradient problem via jump and residual connections. Therefore, ResNet's BasicBlock is utilized as the convolutional block in the encoder of the SANet. The encoder includes a Stom block, six layers, and an APPPM that combines contextual features. Each layer consists of several BasicBlocks arranged in series, except for Layer 6, a bottleneck layer. Studies show that maintaining spatial information can improve segmentation accuracy. The encoder for the SANet network uses a gradual downsampling method in its initial layers, which are Layer 1 through Layer 3. The network has a "buffer" layer after each downsampling layer that preserves detailed information by not performing any downsampling operations. This method helps to maintain high resolution during the encoder's early stage, which allows the network to capture the local details of the image more effectively. It is noteworthy that a higher resolution branch can be employed to preserve spatial details. In the SANet network, a spatial branch (The Dilated Path) is split from Layer 3 to maintain high-resolution features. Unlike previous studies, the spatial branch of SANet incorporates multiple atrous convolutions with varying dilation rates to expand the receptive field. This improves the capacity to discern detailed information in the image. The structure of the DP is shown in Figure~\ref{dp}. After the encoder, the extracted image characteristics are forwarded to the APPPM to complete the integration of information across various scales and sub-regions.

\begin{figure}[H]
\centerline{
\includegraphics[scale=0.08]{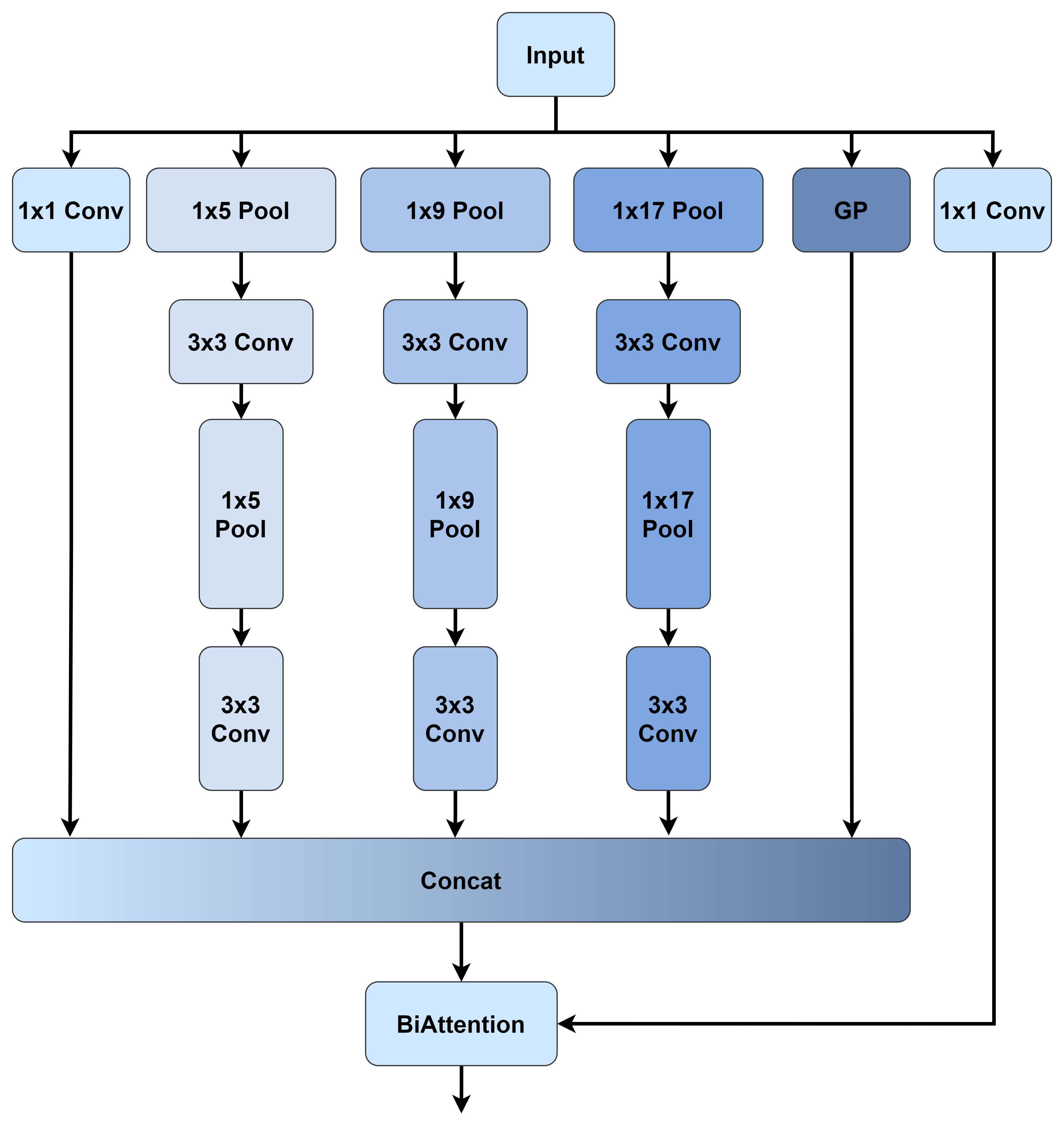}}
\caption{The structures of APPPM. Asymmetric pooling layer provides feature maps with more resolution}
\label{apppm_}
\end{figure}

While the decoder can create semantic predictions at the pixel level using abstract semantic data, the model's prediction speed diminishes with the increased use of decoder layers. Consequently, SANet includes only one carefully designed decoder, SAD, to recover low-resolution features extracted by the encoder. A segmentation header at the end of the network outputs the segmentation results, and in the case of the training phase, an auxiliary segmentation header outputs the segmentation results used for auxiliary losses (including boundary loss and auxiliary semantic loss)

\subsection{\textbf{Asymmetric Pooling Pyramid Pooling Module}}

The pyramid pooling module usually employs different pooling layers of varying sizes to extract image features concurrently. The extracted features are then merged in the channel dimension to produce a composite feature map with multi-scale information. Some recent studies have investigated modifying the quantity and size of pooling layers and incorporating separate convolutional layers following the pooling layers to enhance the feature extraction process. However, there has been limited research on incorporating pooling layers with varying shapes. Therefore, we introduce an Asymmetric Pooling Pyramid Pooling Module (APPPM) as depicted in Figure~\ref{apppm_}.

APPPM consists of three pooling layers with different sizes, including a global pooling layer, except for the fact that the pooling layers have asymmetric shapes. Moreover, we adopted the technique of adding a convolutional layer after the pooling layer, as used in DDRNet. Additionally, we applied channelization to the 1/64 image feature maps via 1$\times$1 convolution to preserve fine details. The resulting feature maps were concatenated as residuals and added to the multi-scale features fusion process. We used the Concat operation to merge the various scale features, which were then passed through the dual-head attention module with the 1/64 image feature map to generate the final output. It is worth noting that the asymmetric pooling layer, which APPPM employs, provides a more in-depth analysis of images with different sizes, allowing for the effortless extraction of their features.

\begin{figure}[H]
\centerline{
\includegraphics[width=3 in]{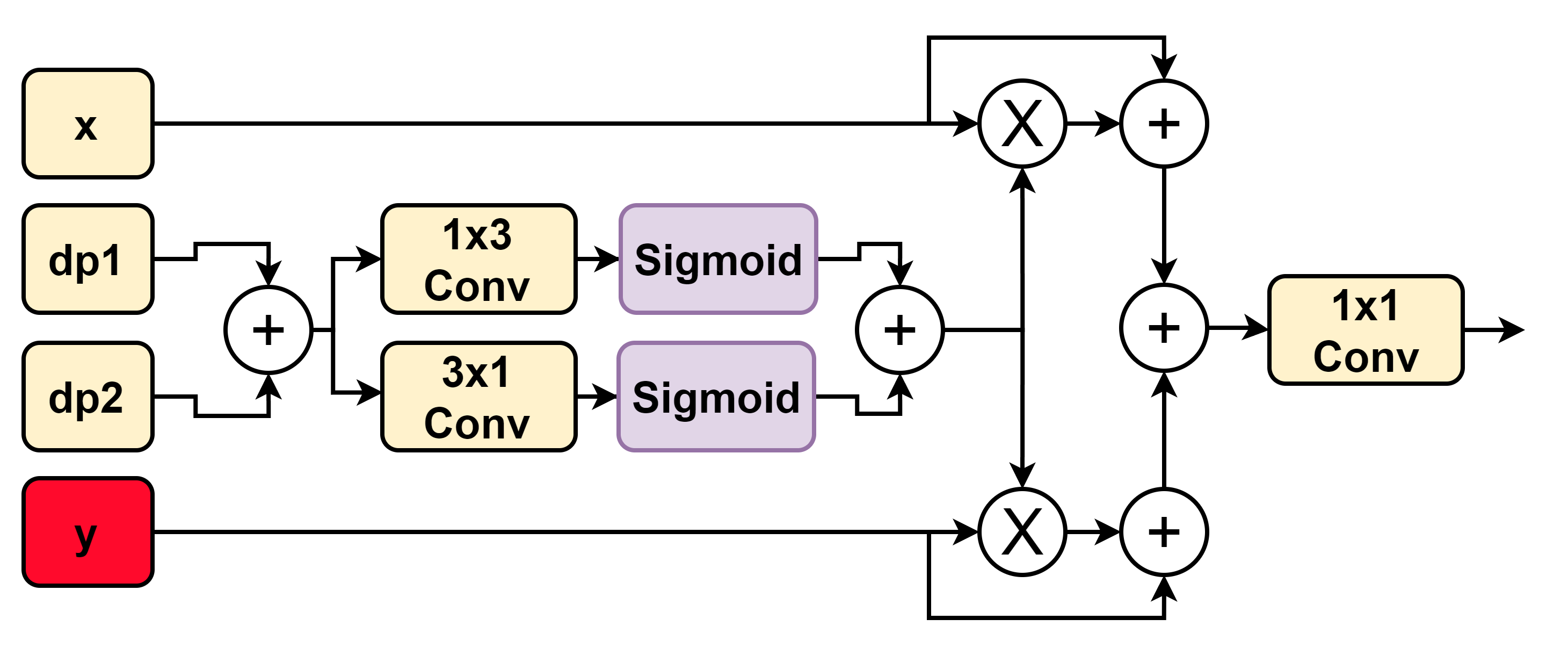}
}
\caption{The implementations of the SAD modules. On the input side, y is the output of APPPM, and the other inputs are the outputs of DP at different times.}
\label{sad}
\end{figure}

\subsection{\textbf{Simple Attention Decoder}}
In the encoder-decoder architecture, the decoder's primary objective is to retrieve the extracted features. The Simple Attention Decoder (SAD) we designed is depicted in Fig~\ref{sad}. The SAD features four inputs: x denotes the final image feature map output from the high-resolution branch. At the same time, dp1 and dp2 correspond to image feature maps from distinct periods of the high-resolution branch, respectively. Additionally, y represents the image feature maps obtained from the high-level semantic features extracted from the image by the APPPM. The dp1 and dp2 undergo an addition operation to unite them into a single feature map, which is then processed through 3$\times$1 and 1$\times$3 asymmetric convolutions to extract horizontal and vertical features, respectively. The resulting transverse and longitudinal features are then utilized in a Sigmoid function to produce an attention mechanism, guiding the fusion of x and y features. Finally, the fused features undergo a 1$\times$1 convolution to adjust the number of channels before being outputted.

If one represents the SAD process as the following formula, where "$\oplus$" represents additive operations, "$\otimes$" represents multiplicative operations, "A" stands for attention, "x" represents high-resolution information, and "y" represents low-resolution information, and $Conv_{1 \times 3}$ and $Conv_{3 \times 1}$ represent asymmetric convolution, then the formula is:
\begin{align}
A_{1} &= Sigmoid(Conv_{1 \times 3}(dp1 \oplus dp2))\\
A_{2} &= Sigmoid(Conv_{3 \times 1}(dp1 \oplus dp2)) \\
A &= A_{1} \oplus A_{2} \\
Output &= Conv_{1\times 1}(x \otimes (1 \oplus A) \oplus y \otimes (1 \oplus (1 - A)))
\end{align}

\section{EXPERIMENTS}

\subsection{\textbf{DataSet}}
\paragraph {\textbf{CityScapes}}
Cityscapes is a substantial dataset for urban segmentation, comprising 5,000 high-resolution images of driving scenes in urban areas, of which there are 20,000 roughly annotated images. Of these, 2,975 high-resolution images were employed for training, 500 for validation, and 1,525 for testing. The images have a resolution of 2048 $\times$ 1024 pixels, and the dataset incorporates 19 categories for semantic segmentation of images. For an equitable comparison with other networks, we utilized only these 5,000 high-resolution images for training.
\paragraph {\textbf{CamVid}}
The Cambridge-driving Labeled Video Database (Camvid) is a dataset that covers road scenes and comprises 701 images of car driving scenes from video sequences. Out of these 701 images, 367 are utilized for training, 101 for validation, and 233 for testing. Each image of this dataset has a resolution of 960$\times$720 pixels, and it consists of a total of 32 categories of labeled information, with 11 categories being employed to execute semantic segmentation tasks.
\subsection{\textbf{Implementation Details}}

\paragraph {\textbf{Pretraining}}
Based on prior research, we commence the training process of SANet by pre-training on the ImageNet\cite{krizhevsky2012imagenet} dataset. This pre-training methodology facilitates faster convergence in image semantic segmentation training for SANet. During the training process, SANet's input resolution was set to 224 $\times$ 224, the batch size to 256, and the number of training epochs to 100. For model optimization, the stochastic gradient descent (SGD) optimizer was employed with a weight decay of 0.0001 and Nesterov momentum of 0.9. Additionally, the model's initial learning rate was set at 0.1 and decreased by one-tenth of the original rate every 30 epochs.

\begin{table*}[ht]
\caption{Comparison of SANet with state-of-the-art real-time segmentation methods on Cityscapes. Training and inference settings are referenced for implementation details.}
\begin{center}
\renewcommand\arraystretch{1.3}
\label{table4}
\begin{tabular}{lcccccccc}
\hline
\multirow{2}{*}{Model} & \multirow{2}{*}{Resolution} & \multirow{2}{*}{FPS(Py.)} & \multirow{2}{*}{FPS(TRT)} &\multirow{2}{*}{GPU} & \multicolumn{2}{c}{mIOU}      & \multirow{2}{*}{Params} & \multirow{2}{*}{Year} \\ \cline{6-7}
                       &                             &                         &                      &                      & val           & test          &        &      \\ \hline
ICNet\cite{zhao2018icnet}                 & 1024$\times$2048                 & 30                    & -                   & TitanX M             & -             & 69.5          & 26.5   & 2017 \\
BiSeNet\cite{yu2018bisenet}               & 768$\times$1536                  & 105.8                    & -                & GTX1080Ti            & 69.0          & 68.4          & 5.8    & 2018 \\ \hline
MSFNet\cite{si2019real}      & 1024$\times$2048                 & 53                    & -                   & RTX2080Ti            & -             & 77.1          & -      & 2019 \\
SwiftNet(RN-18)\cite{orsic2019defense}        & 1024$\times$2048                 & 39.9                     & -                & GTX1080Ti            & 75.5          & 75.4          & 11.8   & 2019 \\
BiSeNetV2\cite{yu2021bisenet}              & 512$\times$1024                  & -                    & 156                  & GTX1080Ti            & 73.4          & 72.6          & -      & 2020 \\

BiSeNetV2-L\cite{yu2021bisenet}            & 512$\times$1024                  & -                   & 47.3                 & GTX1080Ti            & 75.8          & 75.3          & -      & 2020 \\
SFNet(ResNet-18)\cite{li2020semantic}       & 1024$\times$2048                 & 18                     & -                   & GTX1080Ti            & -             & 78.9          & 12.87  & 2020 \\ \hline
CABiNet\cite{yang2021real}                & 1024$\times$2048                 & -                      & 76.5                 & RTX2080Ti            & 76.6          & 75.9          & 2.64   & 2021 \\
STDC1-Seg75\cite{fan2021rethinking}            & 768$\times$1536                  & 126.7                       & -                & GTX1080Ti            & 74.5          & 75.3          & -      & 2021 \\
STDC2-Seg75\cite{fan2021rethinking}            & 768$\times$1536                  & 97                       & -                   & GTX1080Ti            & 77.0          & 76.8          & -      & 2021 \\
LBN-FFN\cite{dong2020real}                & 448$\times$896                   & 51                    & -                   & Titan X              & 74.4          & 73.6          & 6.2    & 2021 \\

MFNet\cite{lu2022mfnet}                  & 512$\times$1024                  & 116                       & -                  & Tian XP              & -             & 72.1          & 1.34   & 2022 \\
DMANet\cite{weng2022deep}                 & 1024$\times$2048                 & 46.7                    & -                 & GTX1080Ti            & 77.4          & 77.0          & 14.6   & 2022 \\ 
PP-LiteSeg-B2\cite{lu2022mfnet}                  & 768$\times$1536                  & -                       & 102.6                 & GTX1080Ti               & 78.2             & 77.5          & -   & 2022 \\
HoloParser-34\cite{li2022holoparser}          & 1024$\times$2048                 &  81                         & 180                 & RTX2080Ti            & -          & 77.26          & \textbf{-}   & -    \\
DMRNet-L\cite{wang2023deep}            & 1024$\times$2048                 &  68.7                       & -                 & RTX2080Ti            & 78.2 & 77.6 & 6.93   & 2023    \\ 
\hline
DDRNet-S\cite{pan2022deep}                 & 1024$\times$2048                 & 101.6                    & -                & RTX2080Ti            & 77.8 /  79.3         & 76.4 / 77.1          & 5.7    & 2022 \\
FASSD-Net\cite{lu2022mfnet}                  & 1024$\times$2048                  & 41.1                       & -                 & GTX1080Ti               & 78.8 / 79.7             & 76.0 / 77.5          & 2.85   & 2022 \\
PIDNet-S\cite{xu2023pidnet}                  & 1024$\times$2048                  & 90                       & -                 & RTX2080Ti               & 78.8 / 79.9             & 76.9 / 78.1          & 7.6   & 2023 \\
\textbf{SANet-S}               & 1024$\times$2048                 &  65                       & -                 & RTX2080Ti            & 78.6 / \textbf{79.9} & \textbf{77.2} / \textbf{78.4} & 8.25   & -    \\ 
\textbf{SANet-M}               & 1024$\times$2048                 &  52                       & -                 & RTX2080Ti            & 78.8 / \textbf{80.2} & \textbf{77.6} / \textbf{78.8} & 11.98   & -    \\ 
\hline

\end{tabular}
\end{center}
\end{table*}

\paragraph {\textbf{Train}}
To ensure objectivity, a similar training procedure as applied in previous work was employed for fairness. We compared two learning rate strategies, namely cosine annealing and the "poly" strategy, and ultimately selected the "poly" strategy for its more effective model accuracy improvement. The power in the "poly" strategy was set to 0.9. To improve the data, we utilize standardized methods, like random scaling and cropping, with parameters ranging from 0.5 to 2.0. Furthermore, we offer alternatives for other prevalent hyperparameters, including training epochs of 484 (equivalent to around 120,000 iterations) and 150, batch sizes of 12 and 8, initial learning rates of 0.01 and 0.001, weight decay of 0.0005 and 0.0005, image cropping sizes of 1024 × 1024 and 960 $\times$ 720.

In addition, we attempted to fine-tune the standard hyperparameters at Cityscapes. However, we observed a minor enhancement in the model's performance. In regards to the training epoch, we endeavored to adjust it to 324 (equivalent to 60,000 iterations, according to STDC) and also attempted to increase it to 500. With regards to batch size, our initial setting was 8 (approximately 180,000 iterations). We noted that a batch size of 8 outperformed a batch size of 12 when ImageNet pre-training was not employed. However, with the use of ImageNet pre-training, the results were inverted, and we assume that overfitting may be implicated. Regarding the tuning of the learning rate, we attempted to use [0.02, 0.015, 0.005, 0.001] as the initial learning rate. However, we discovered that a learning rate greater than 0.01 typically results in unstable findings, while a learning rate below 0.01 may result in early convergence.
\paragraph {\textbf{Inference}}
In real-time semantic segmentation, it is common to use the 2080Ti GPU to test the inference speed of models. To conduct our inference experiments, we employed a singular 2080Ti GPU and an experimental environment encompassing Anaconda, Pytorch 1.12, CUDA 11.3, and cuDNN 8.0. The protocol we followed was identical to that of prior research~\cite{peng2022pp}~\cite{xu2023pidnet}. For the CityScapes and CamVid datasets, we selected input image sizes of 2048 $\times$ 1024 and 960 $\times$ 720, respectively, with a batch size of 1. We eliminated the segmentation header, which was used for auxiliary loss, and incorporated the batch normalization layer into the convolutional layer.

\subsection{\textbf{Ablation Study}}
In the ablation experiment, we assessed every module of SANet. The test settings remained constant. Only the modules affected by the ablation experiments were modified.

\paragraph {\textbf{Network Structure}}
In order to verify the effectiveness of the combination of the individual modules, we trained only the encoder part of SANet on the ImageNet dataset in addition to the APPPM, which consists of a primary branch responsible for downsampling and an atrous convolution branch responsible for expanding the field of view. The trained encoder will be used as a baseline result for comparison, with the primary and atrous convolution branches using addition operations to fuse features. As shown in Table~\ref{network_structure}, the addition operation is a traditional method of feature fusion, but simply adding two feature maps may result in different information at the exact location, canceling each other out and leading to information loss.

\begin{table}[H]
\caption{Ablation experiment of Network Structure. The baseline model pertains to the layers from Stom to Layer 6 and the atrous convolution branch inside SANet. The baseline was pre-trained in the usual manner on the ImageNet dataset.}
\begin{center}
\renewcommand\arraystretch{1.3}
\label{network_structure}
\begin{tabular}{|l|c|c|c|}
\hline
Module                 & Speed(FPS) & Params(M) & mIOU(\%)       \\ \hline
Baseline + add         & 76.5       & 5.7       & 71.2          \\ \hline
Baseline + SAD         & 73.6       & 5.83      & 74.9          \\ \hline
Baseline + APPPM + add & 68.3       & 8.2       & 77.6          \\ \hline
Baseline + APPPM + SAD & 65.1       & 8.26      & \textbf{78.1} \\ \hline
SANet-S                & 65.1       & 8.26      & \textbf{78.6} \\ \hline
\end{tabular}
\end{center}
\end{table}

The introduction of SAD helps to alleviate the information loss problem, with the network being able to capture complex relationships between features and selectively focus on features that are more important to the task. Pyramid pooling is able to capture image information at different scales, and the addition of APPPM can significantly improve the perceptual ability of the model, leading to a rapid increase in network performance. The improvement in accuracy will be more significant when combining APPPM and SAD.

\begin{figure*}[h]

\centerline{
 \begin{minipage}{0.15\linewidth}
 	\vspace{3pt}
 	\centerline{\includegraphics[width=\textwidth]{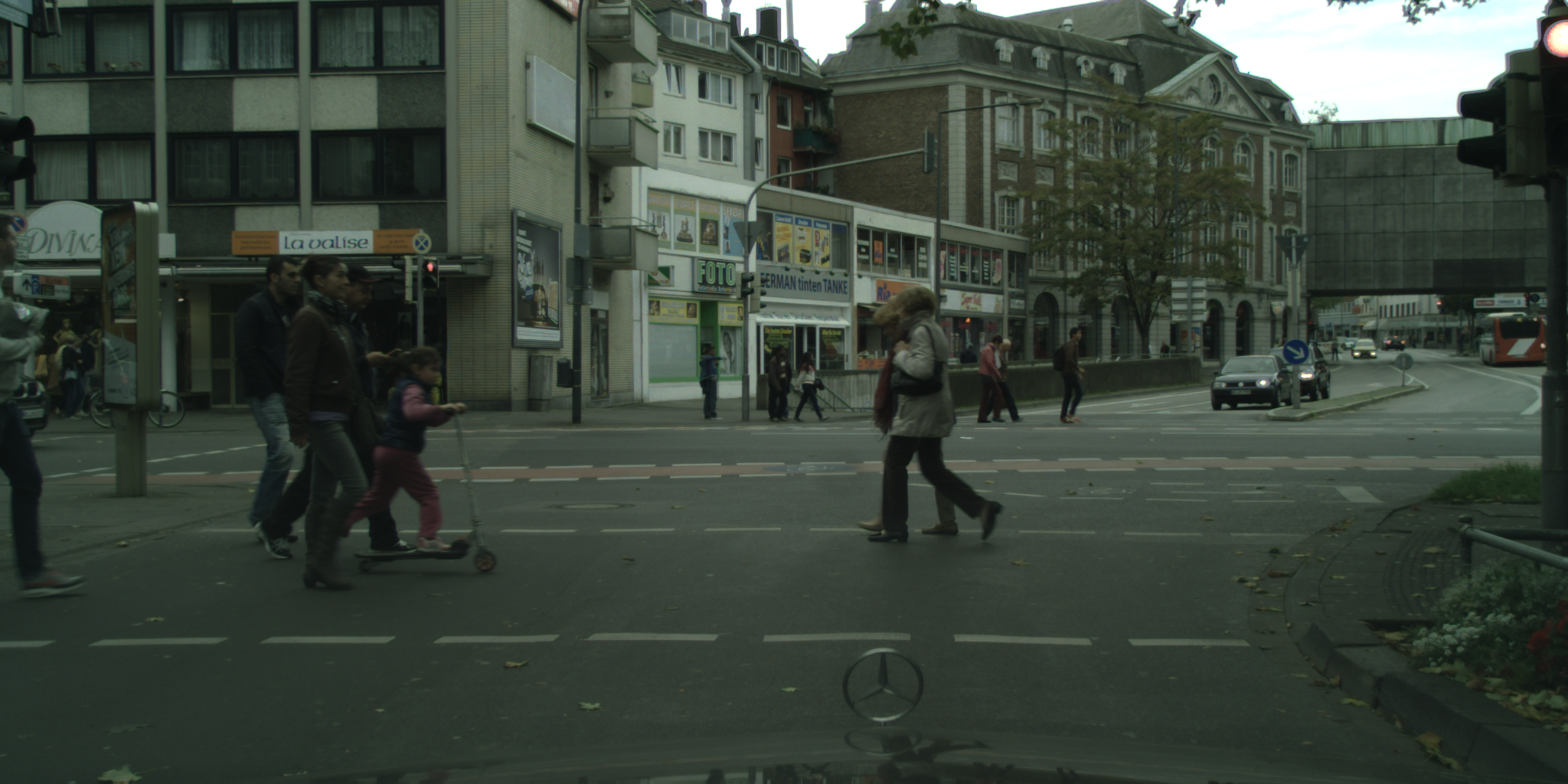}}
 	\vspace{3pt}
 	\centerline{\includegraphics[width=\textwidth]{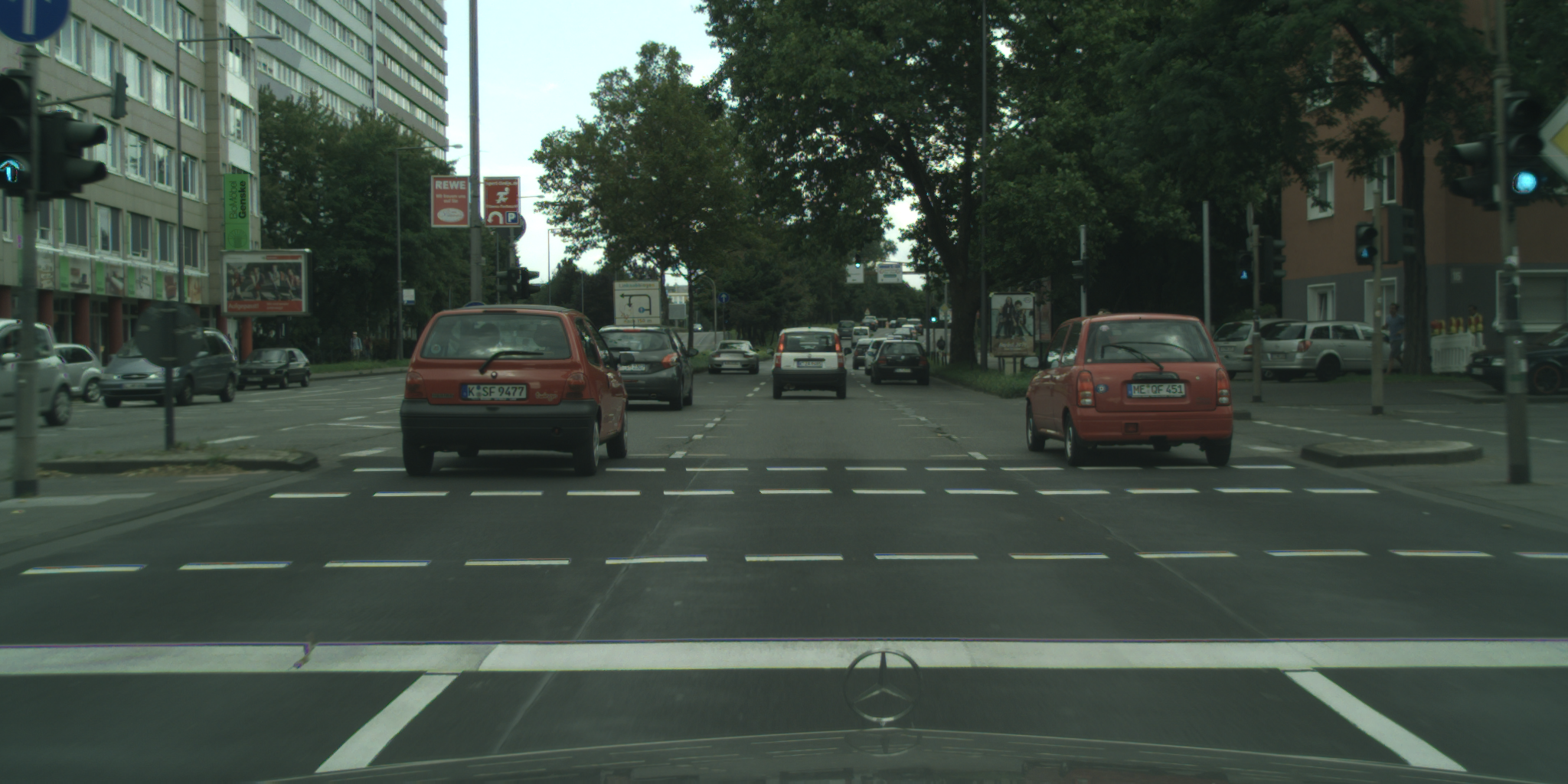}}
 	\vspace{3pt}
 	\centerline{\includegraphics[width=\textwidth]{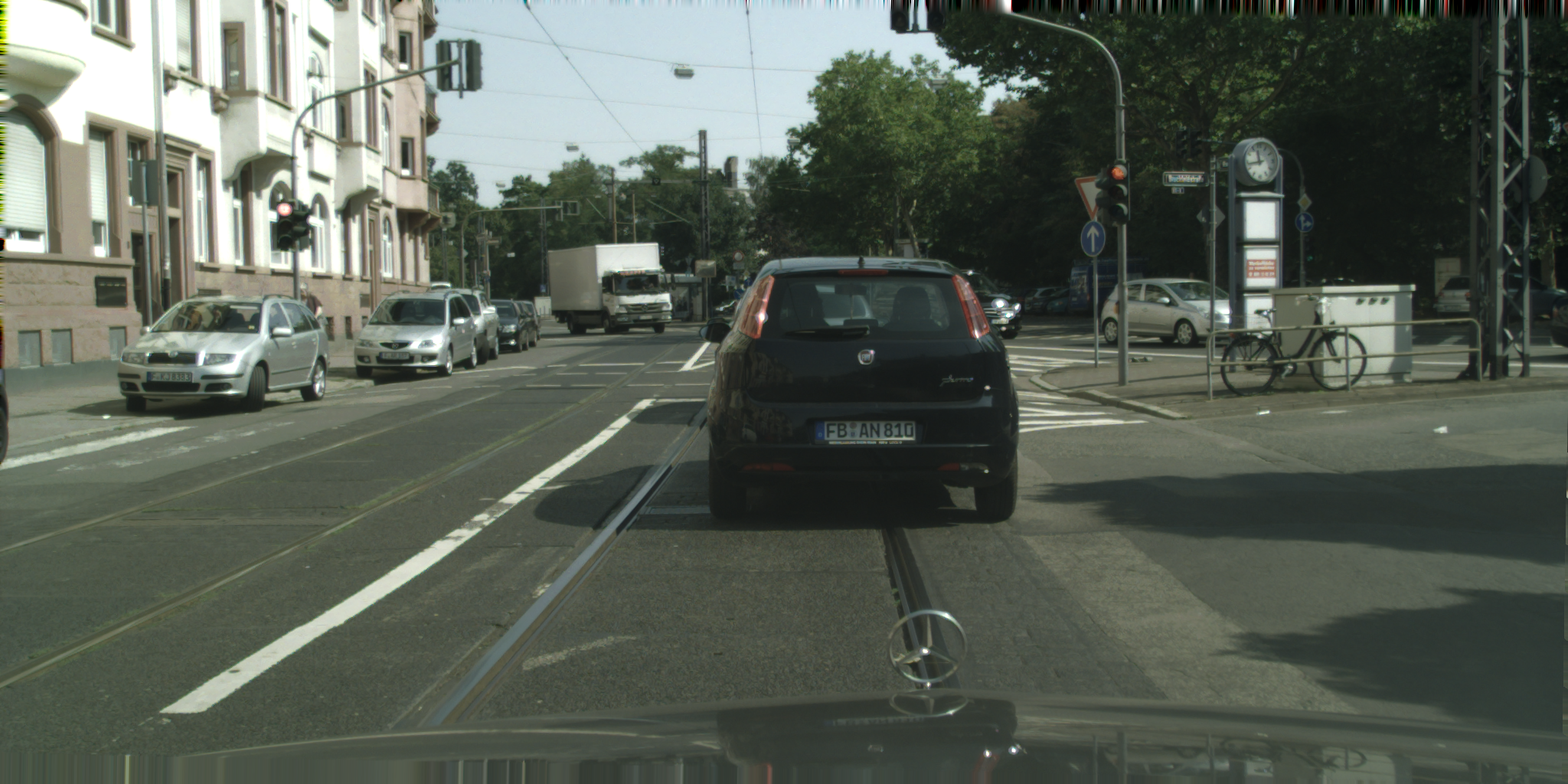}}
 	\vspace{3pt}
 	\centerline{\includegraphics[width=\textwidth]{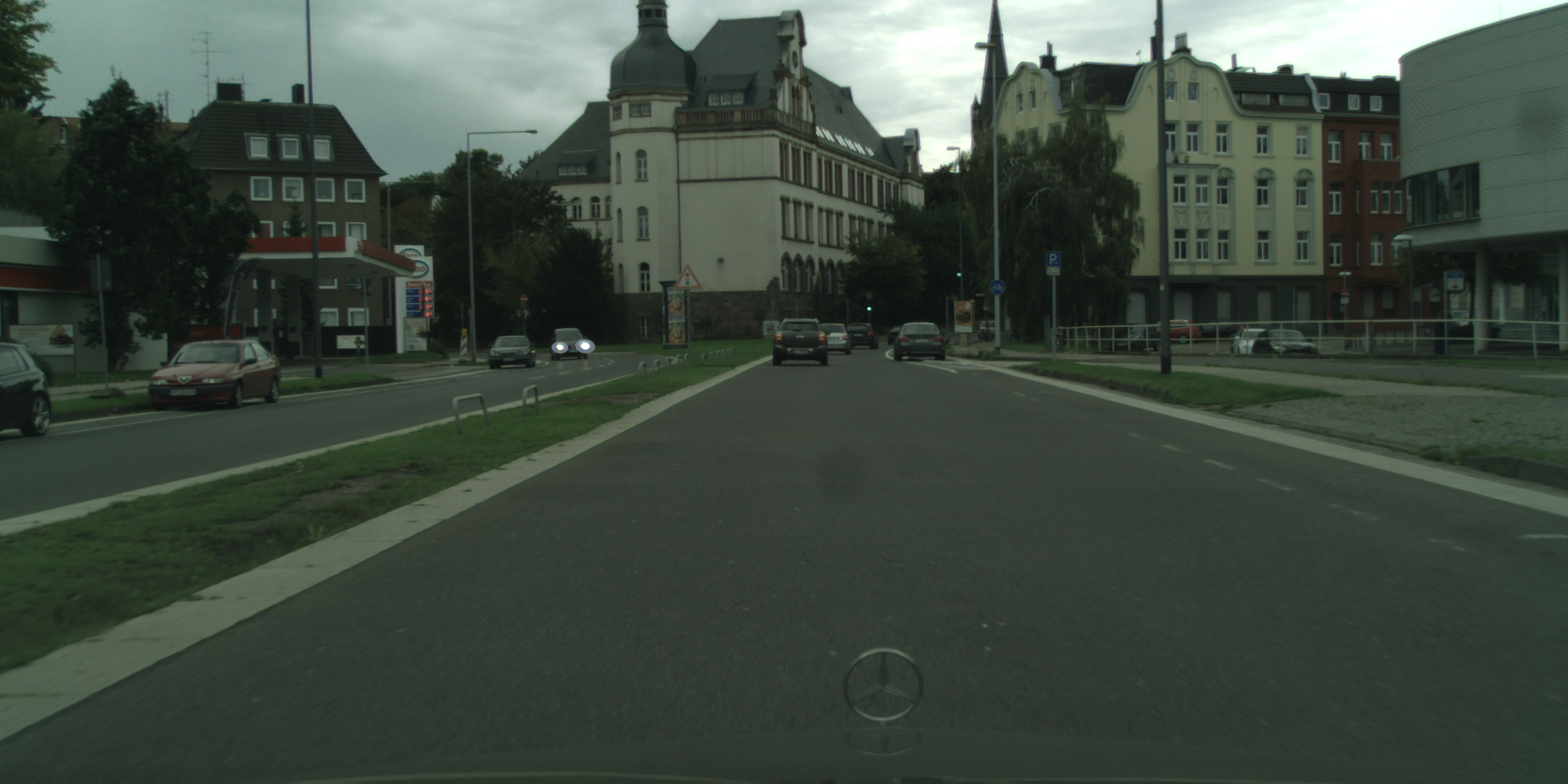}}
 \end{minipage}
 \begin{minipage}{0.15\linewidth}
 	\vspace{3pt}
 	\centerline{\includegraphics[width=\textwidth]{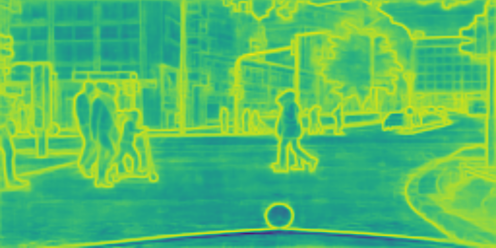}}
 	\vspace{3pt}
 	\centerline{\includegraphics[width=\textwidth]{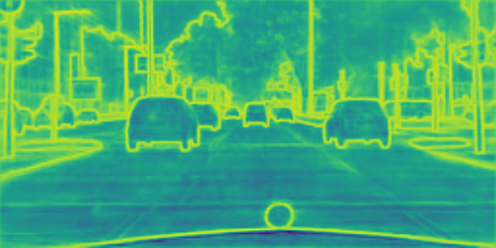}}
 	\vspace{3pt}
 	\centerline{\includegraphics[width=\textwidth]{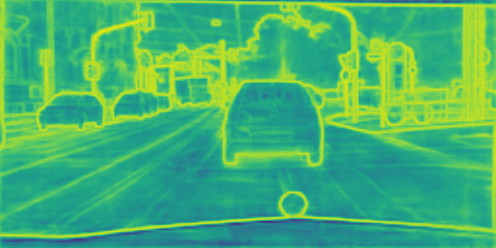}}
 	\vspace{3pt}
 	\centerline{\includegraphics[width=\textwidth]{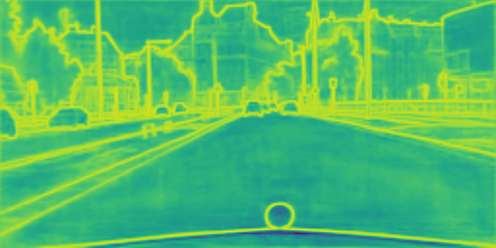}}
 \end{minipage}
 \begin{minipage}{0.15\linewidth}
	\vspace{3pt}
	\centerline{\includegraphics[width=\textwidth]{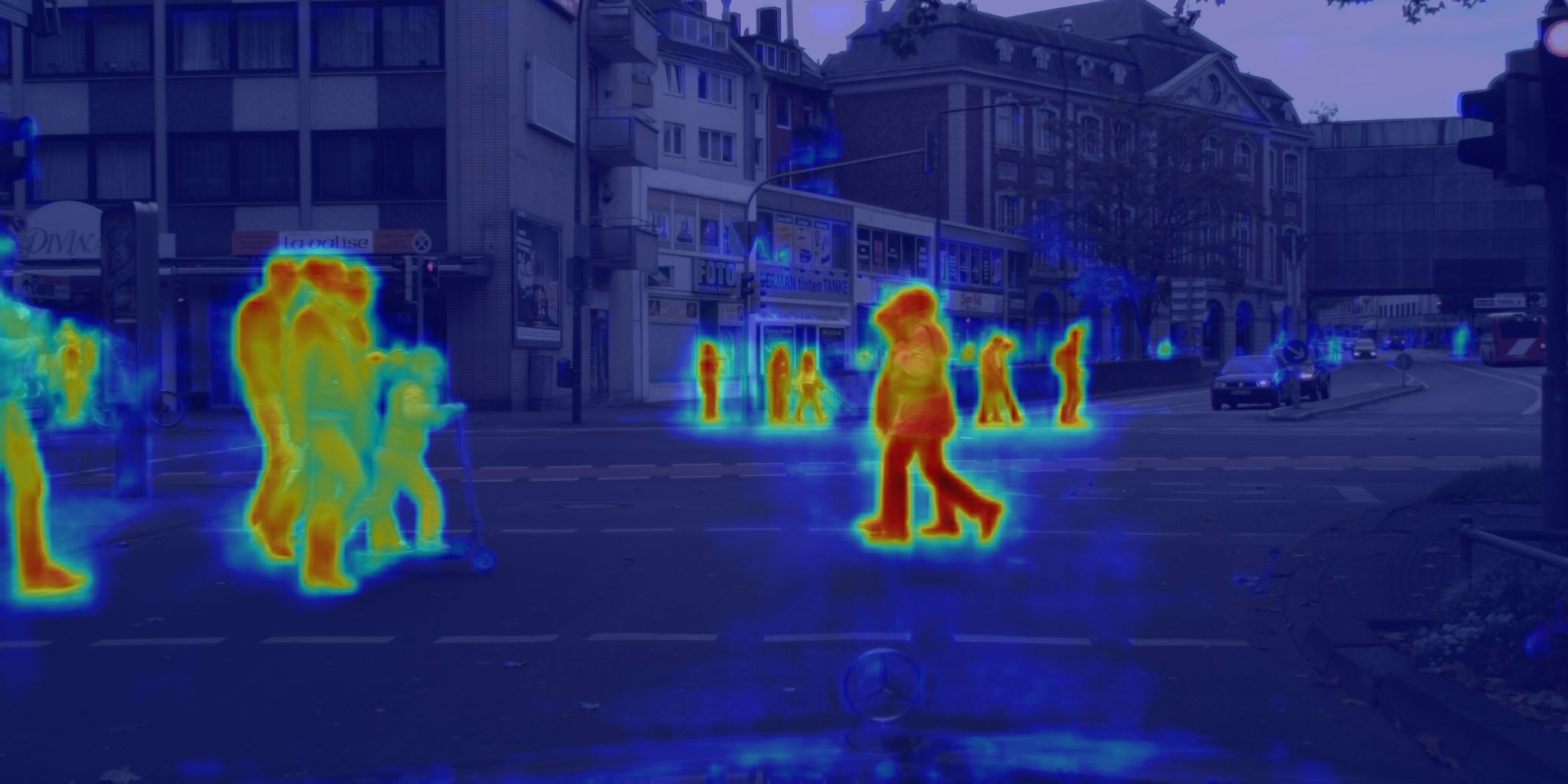}}
	\vspace{3pt}
	\centerline{\includegraphics[width=\textwidth]{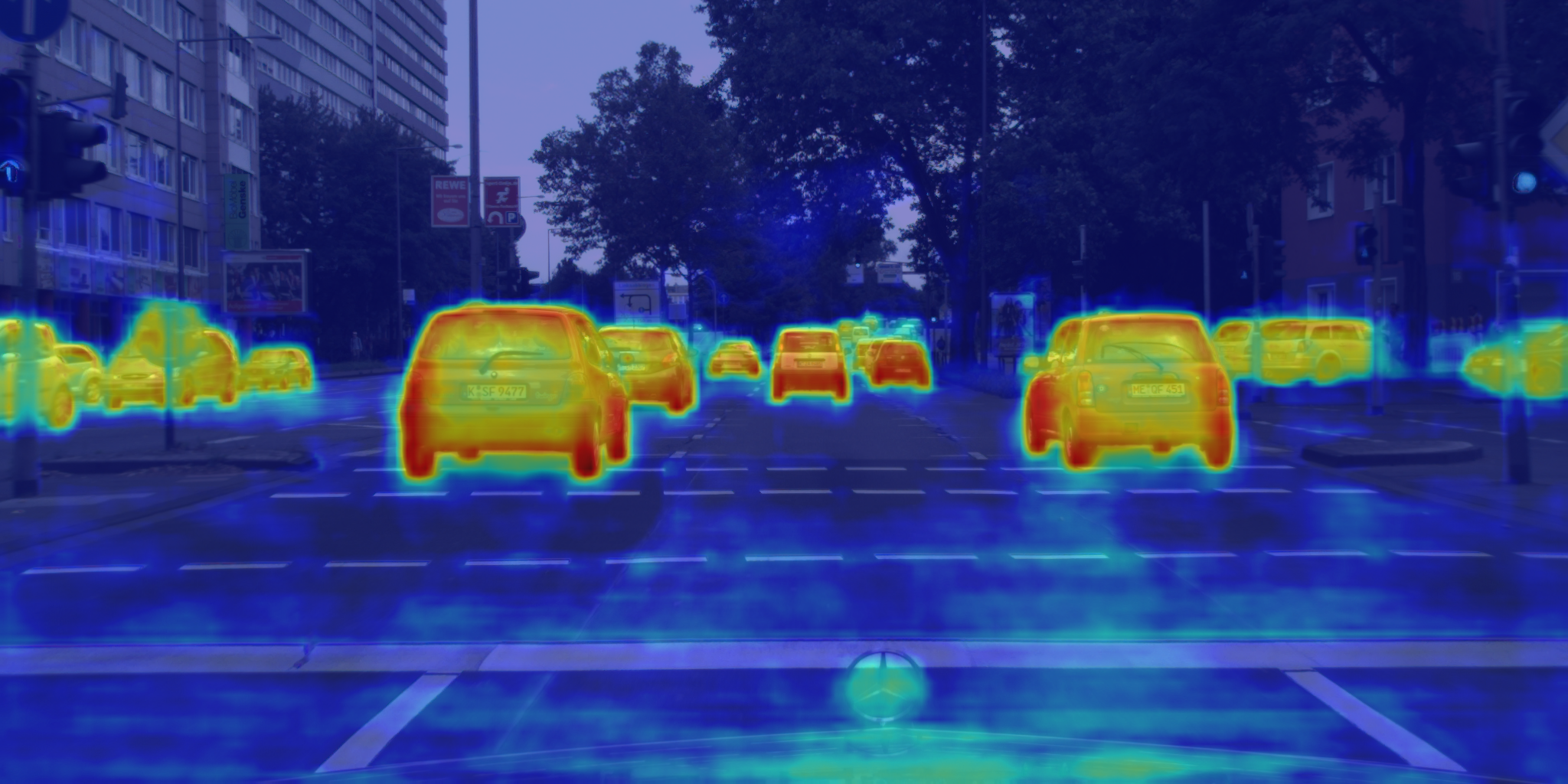}}
	\vspace{3pt}
	\centerline{\includegraphics[width=\textwidth]{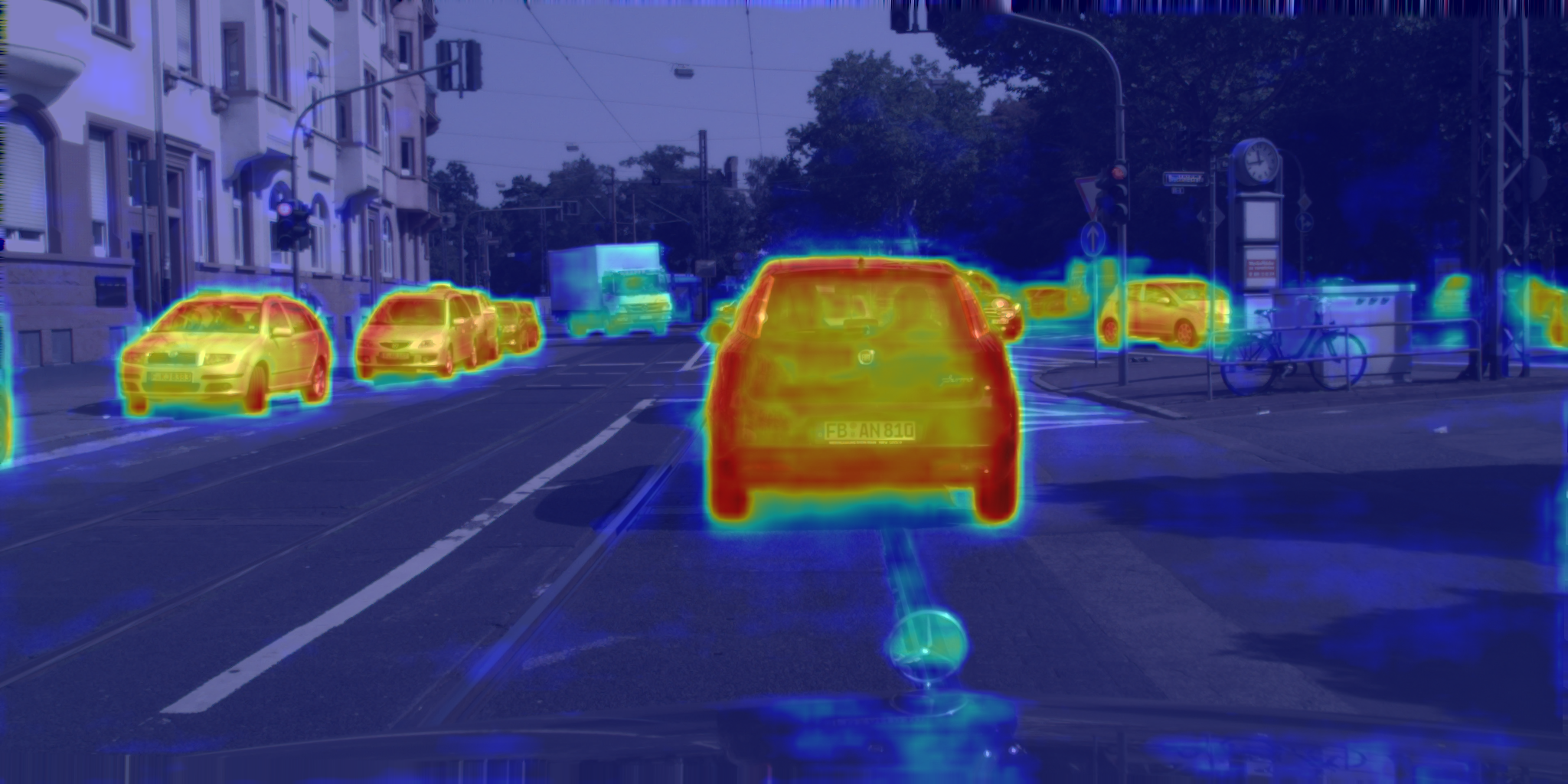}}
	\vspace{3pt}
	\centerline{\includegraphics[width=\textwidth]{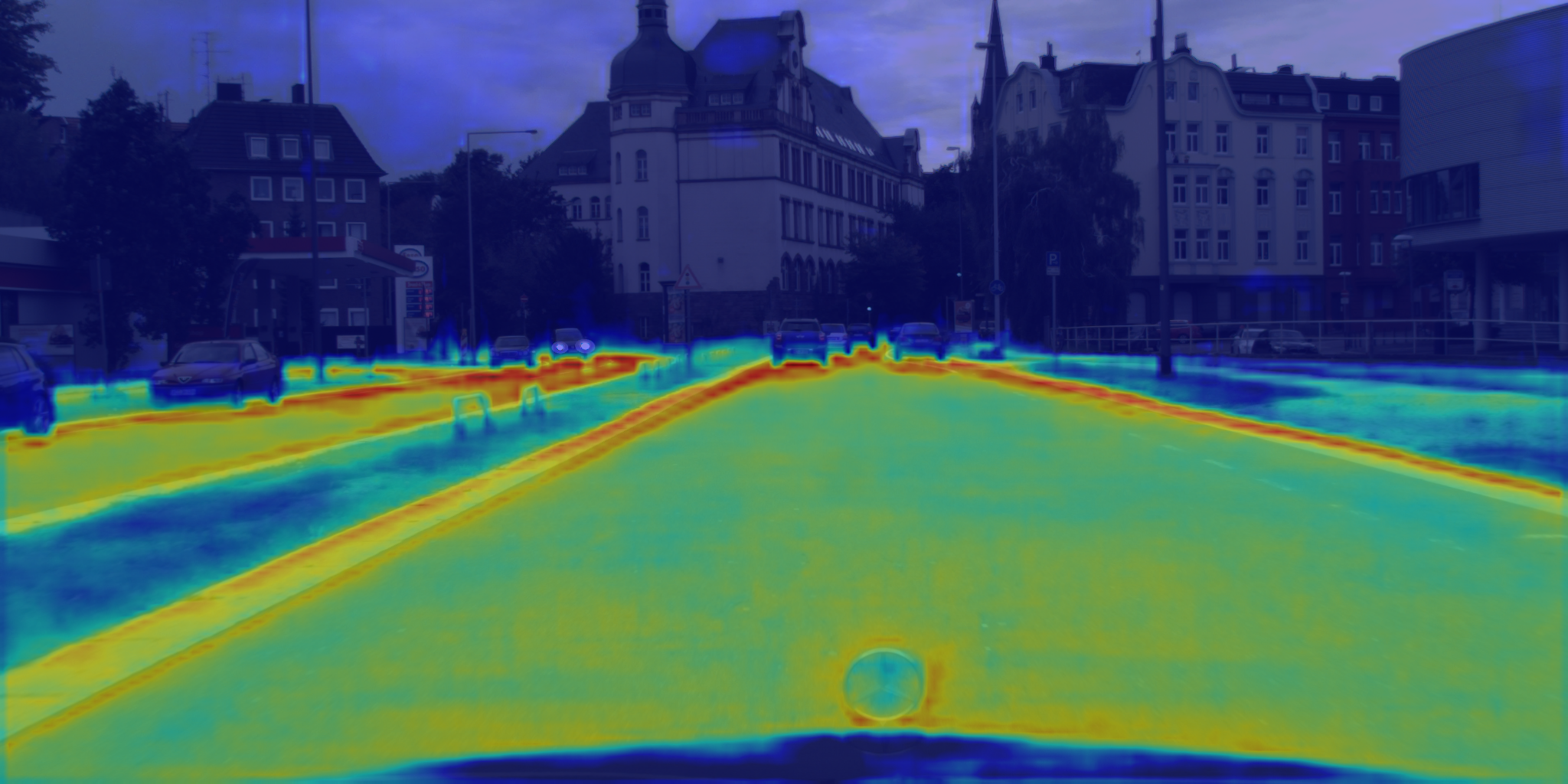}}
\end{minipage}
\begin{minipage}{0.15\linewidth}
	\vspace{3pt}
	\centerline{\includegraphics[width=\textwidth]{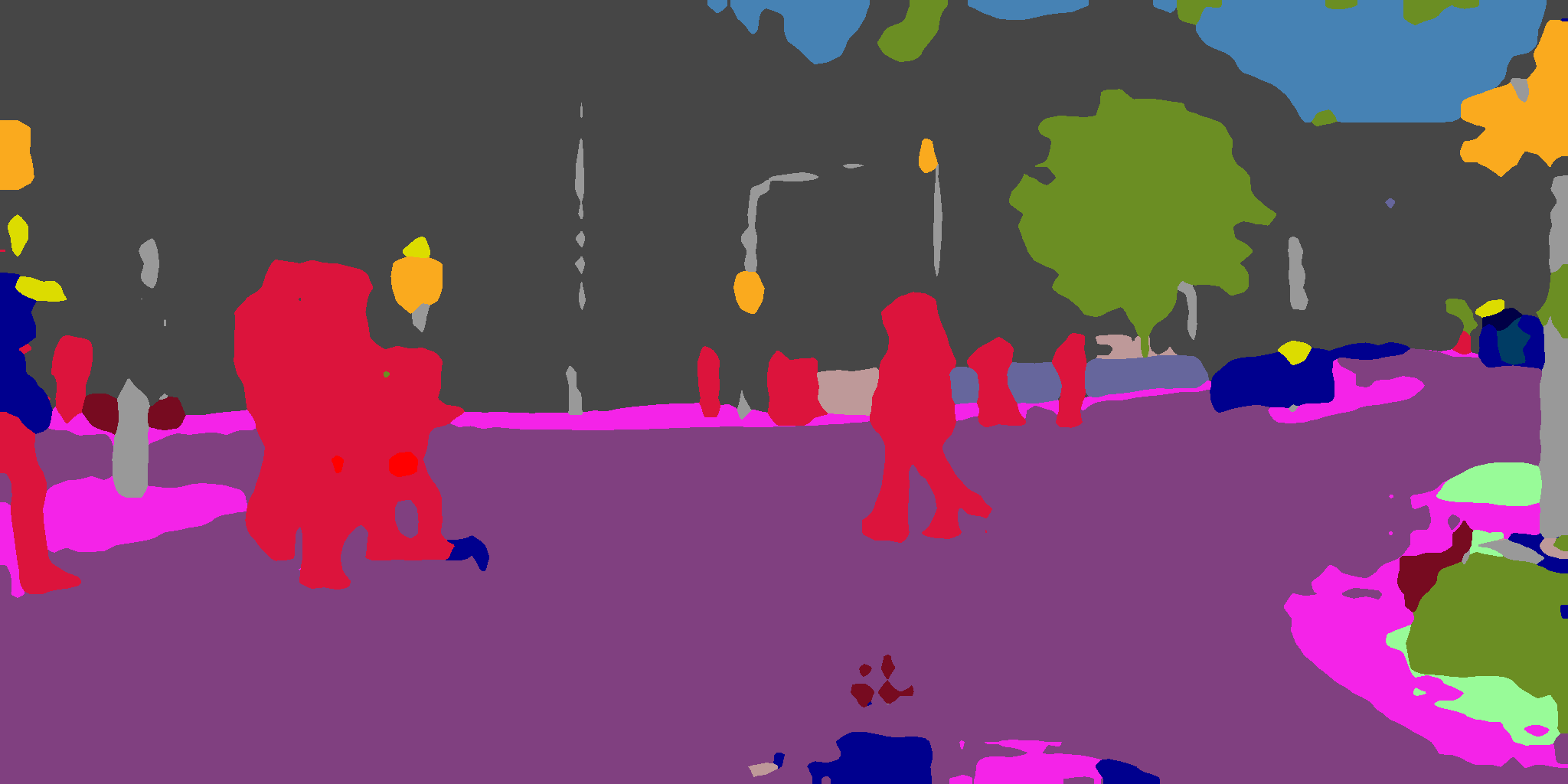}}
	\vspace{3pt}
	\centerline{\includegraphics[width=\textwidth]{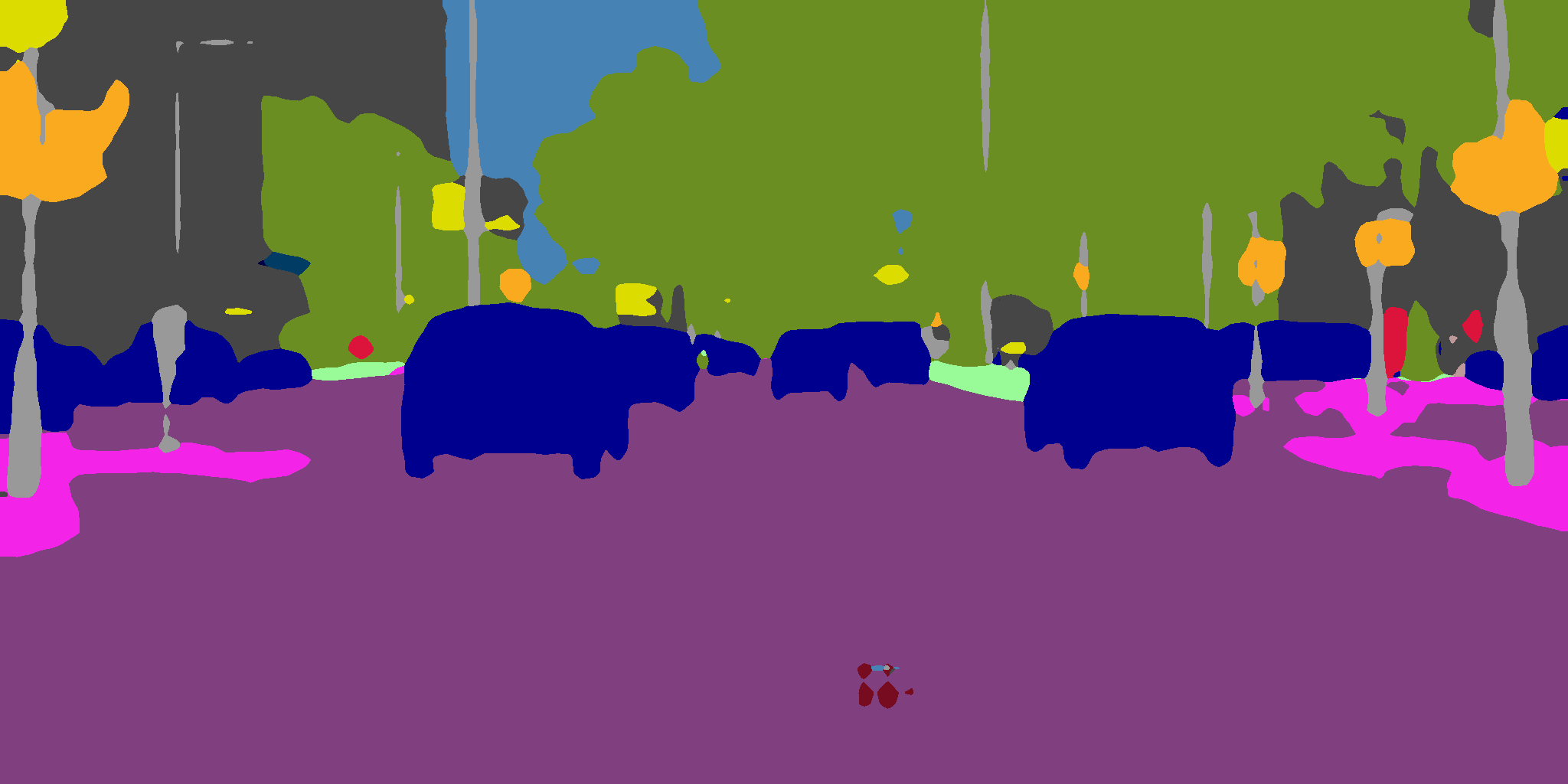}}
	\vspace{3pt}
	\centerline{\includegraphics[width=\textwidth]{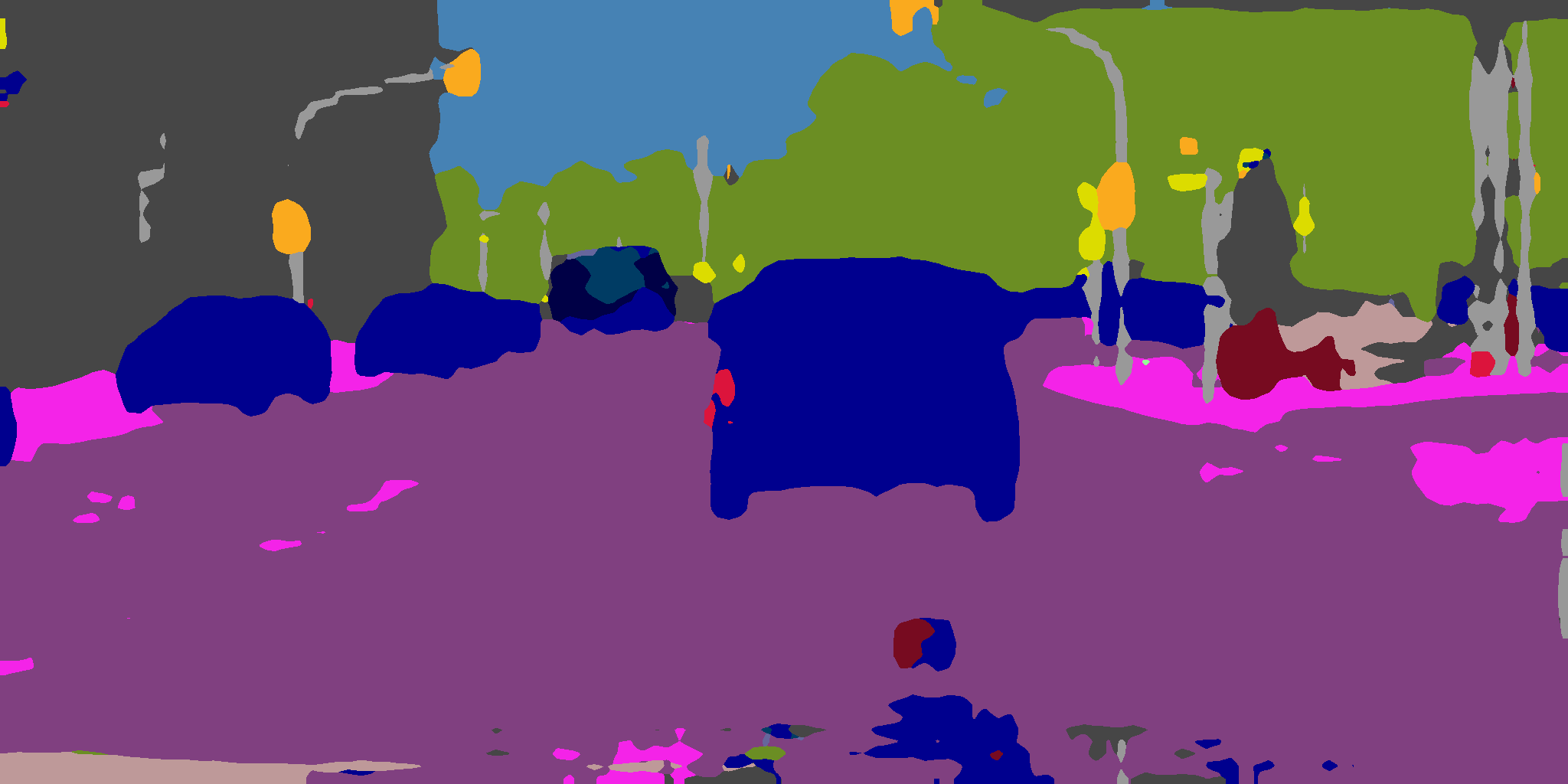}}
	\vspace{3pt}
	\centerline{\includegraphics[width=\textwidth]{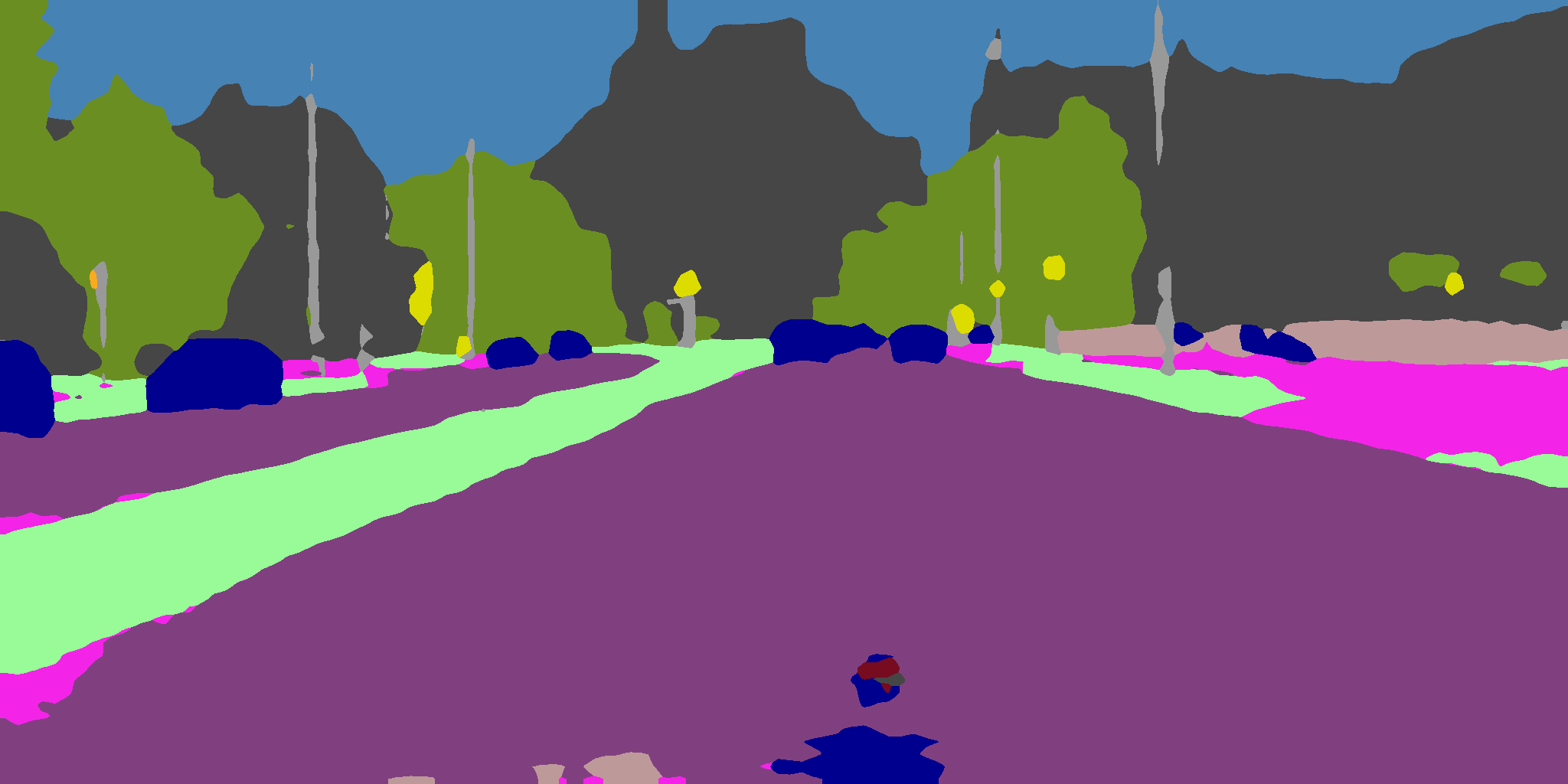}}
\end{minipage}
\begin{minipage}{0.15\linewidth}
	\vspace{3pt}
	\centerline{\includegraphics[width=\textwidth]{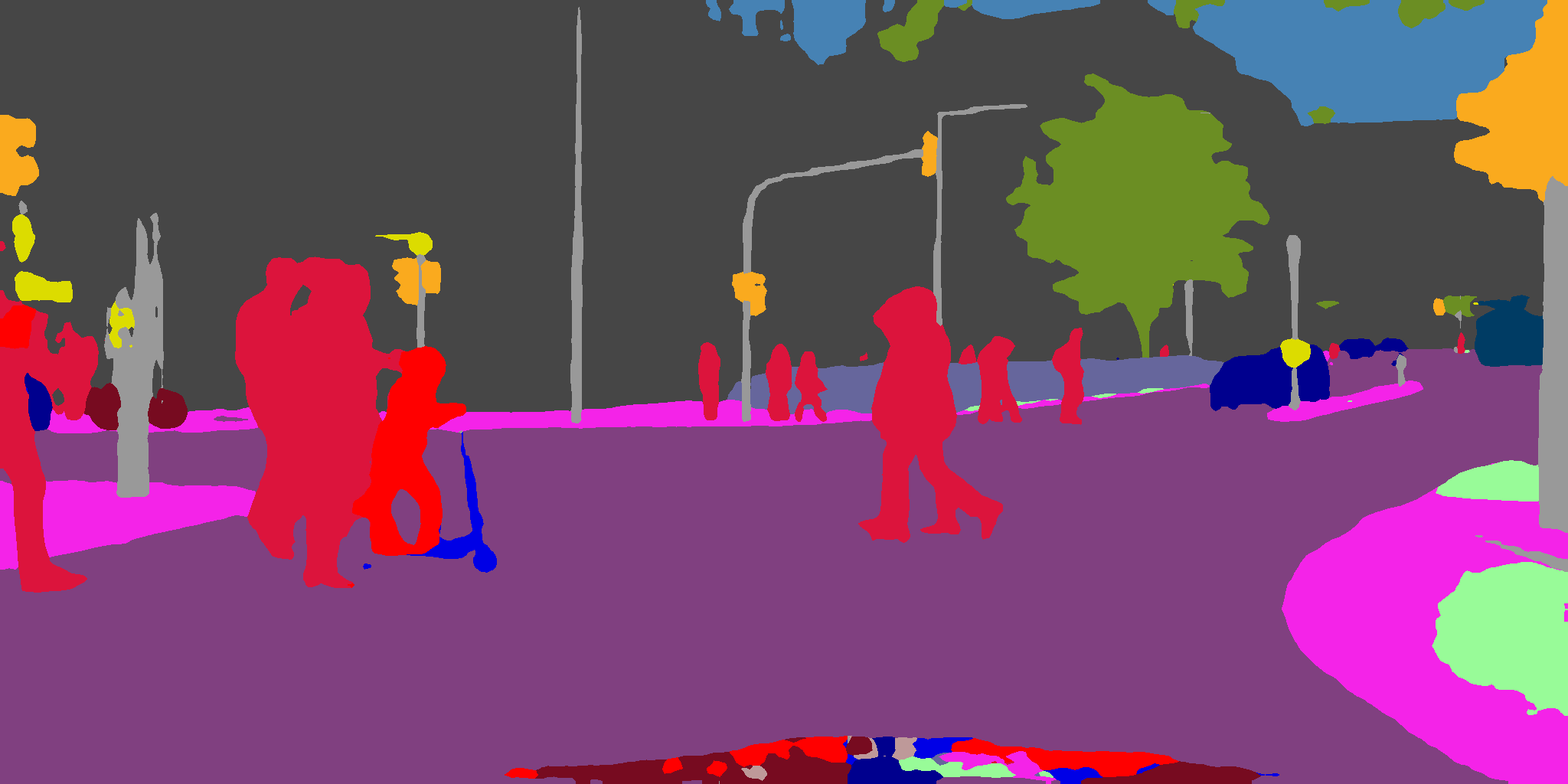}}
	\vspace{3pt}
	\centerline{\includegraphics[width=\textwidth]{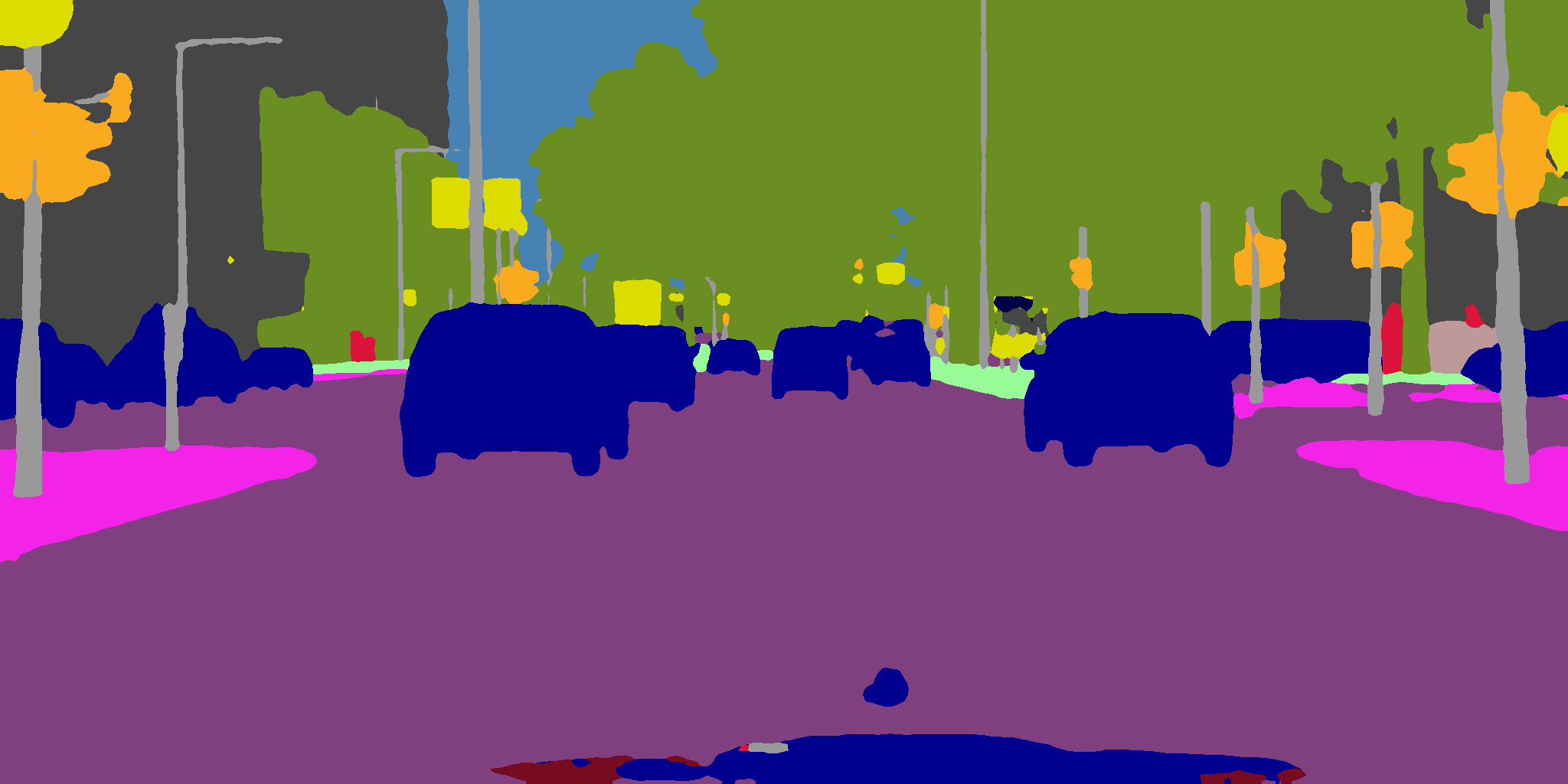}}
	\vspace{3pt}
	\centerline{\includegraphics[width=\textwidth]{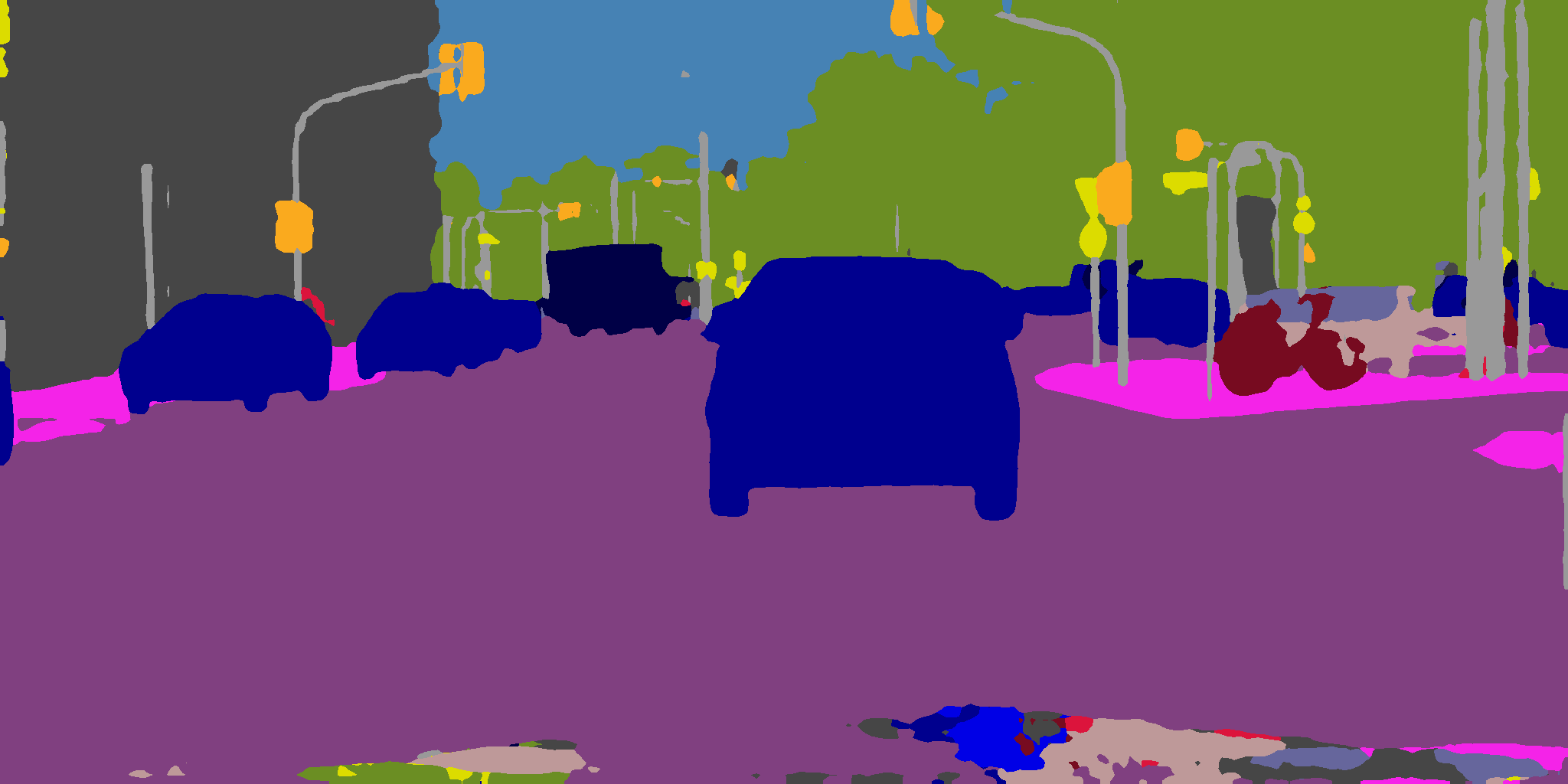}}
	\vspace{3pt}
	\centerline{\includegraphics[width=\textwidth]{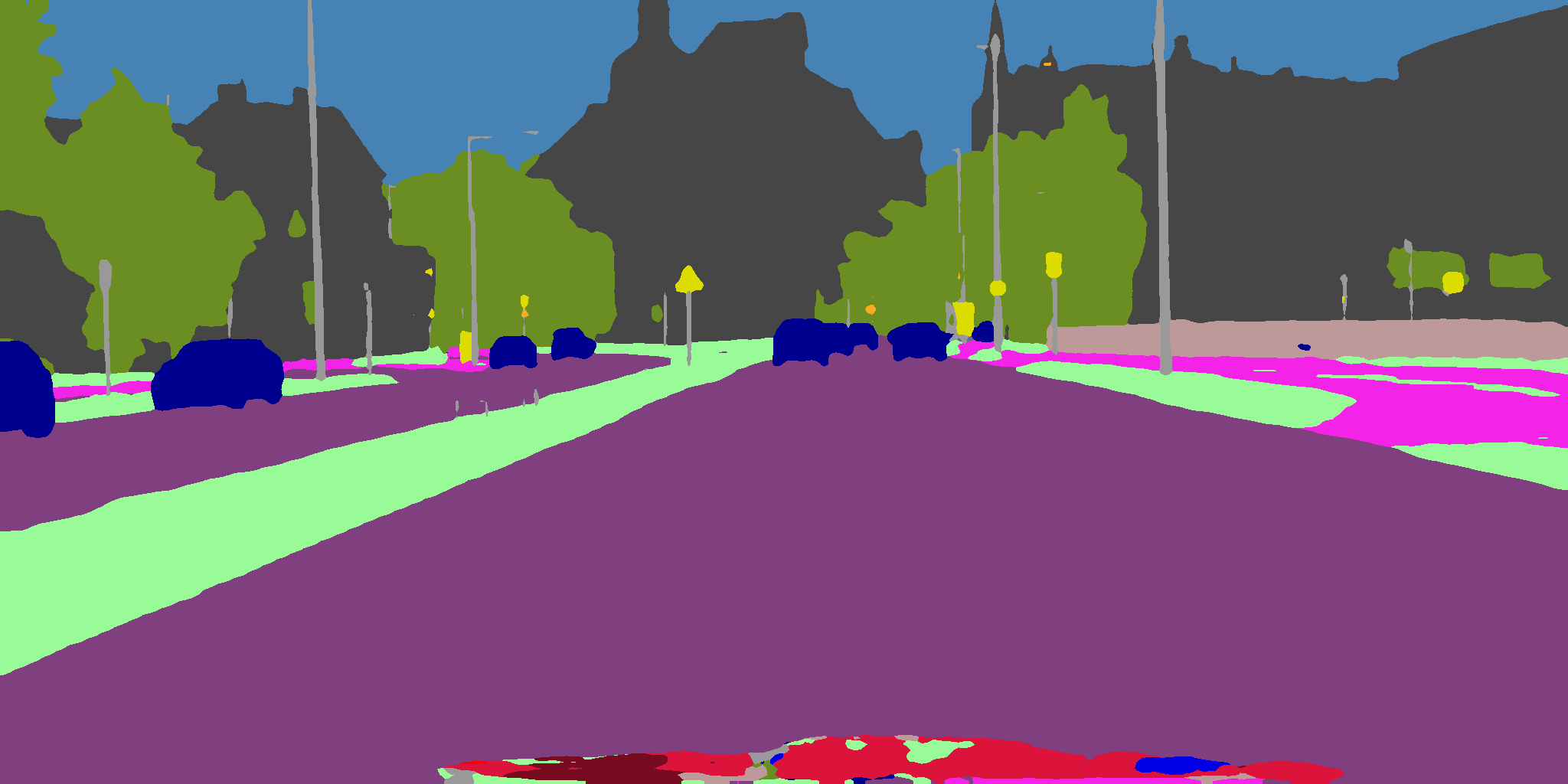}}
\end{minipage}
\begin{minipage}{0.15\linewidth}
	\vspace{3pt}
	\centerline{\includegraphics[width=\textwidth]{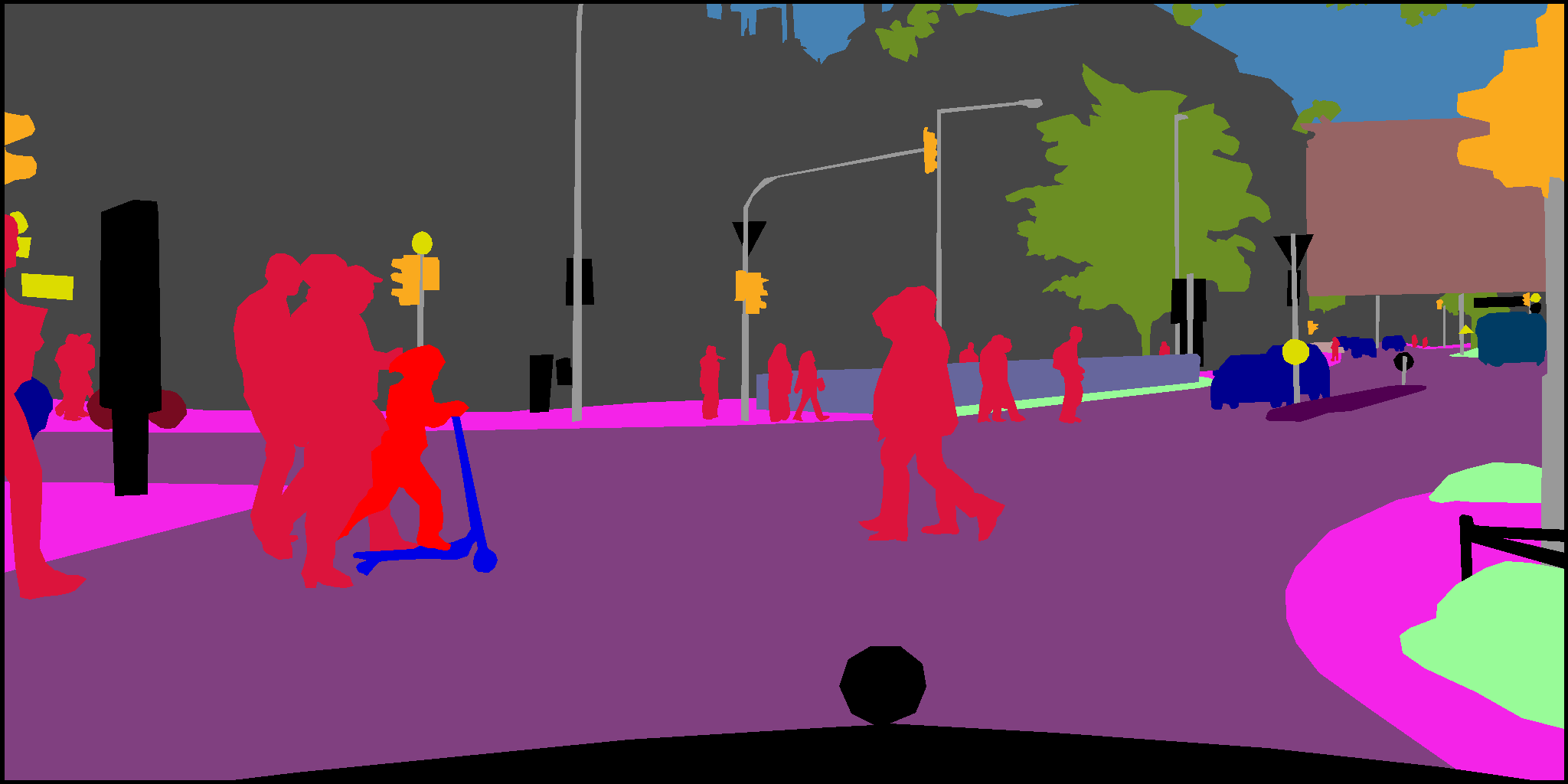}}
	\vspace{3pt}
	\centerline{\includegraphics[width=\textwidth]{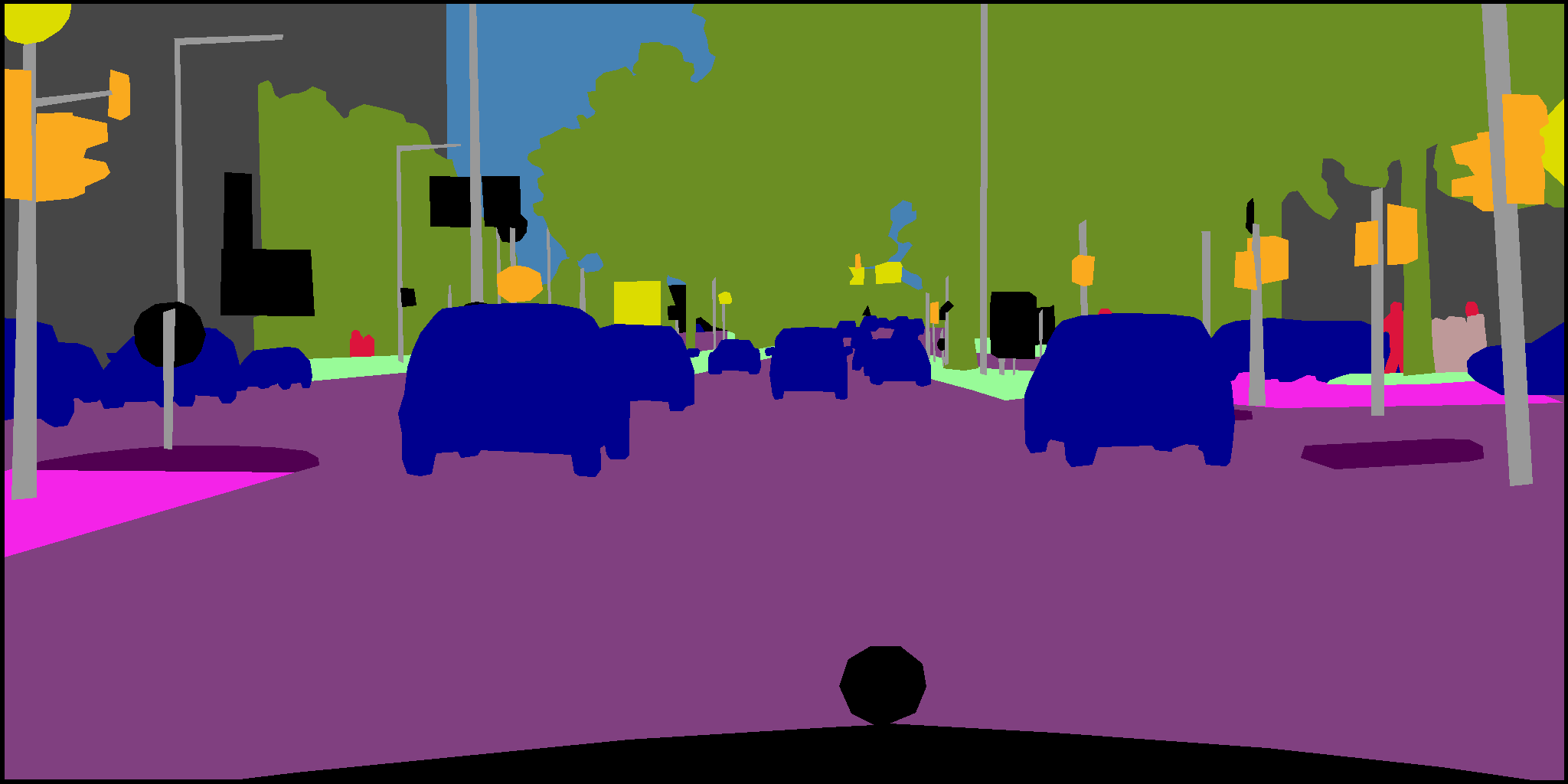}}
	\vspace{3pt}
	\centerline{\includegraphics[width=\textwidth]{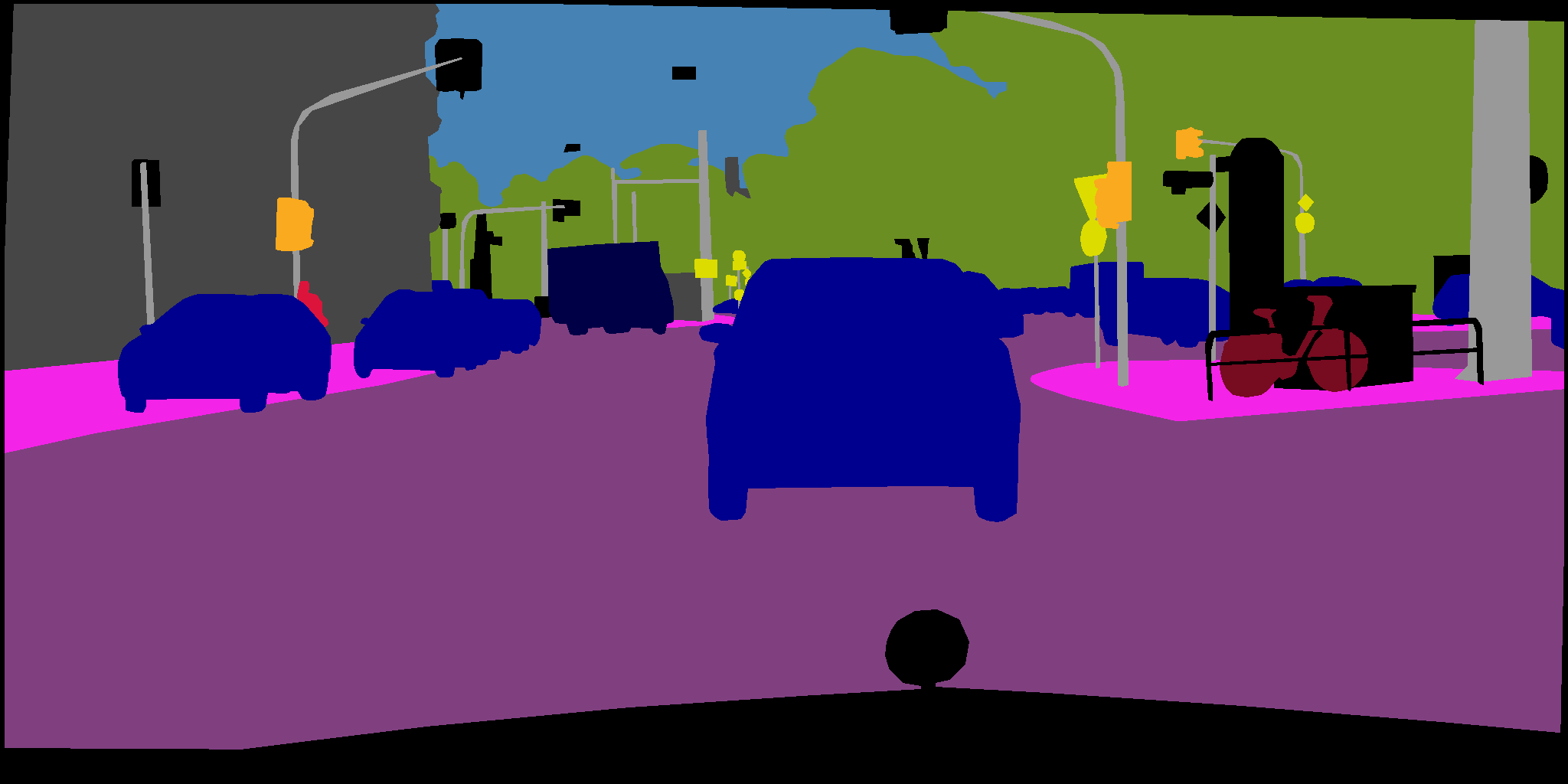}}
	\vspace{3pt}
	\centerline{\includegraphics[width=\textwidth]{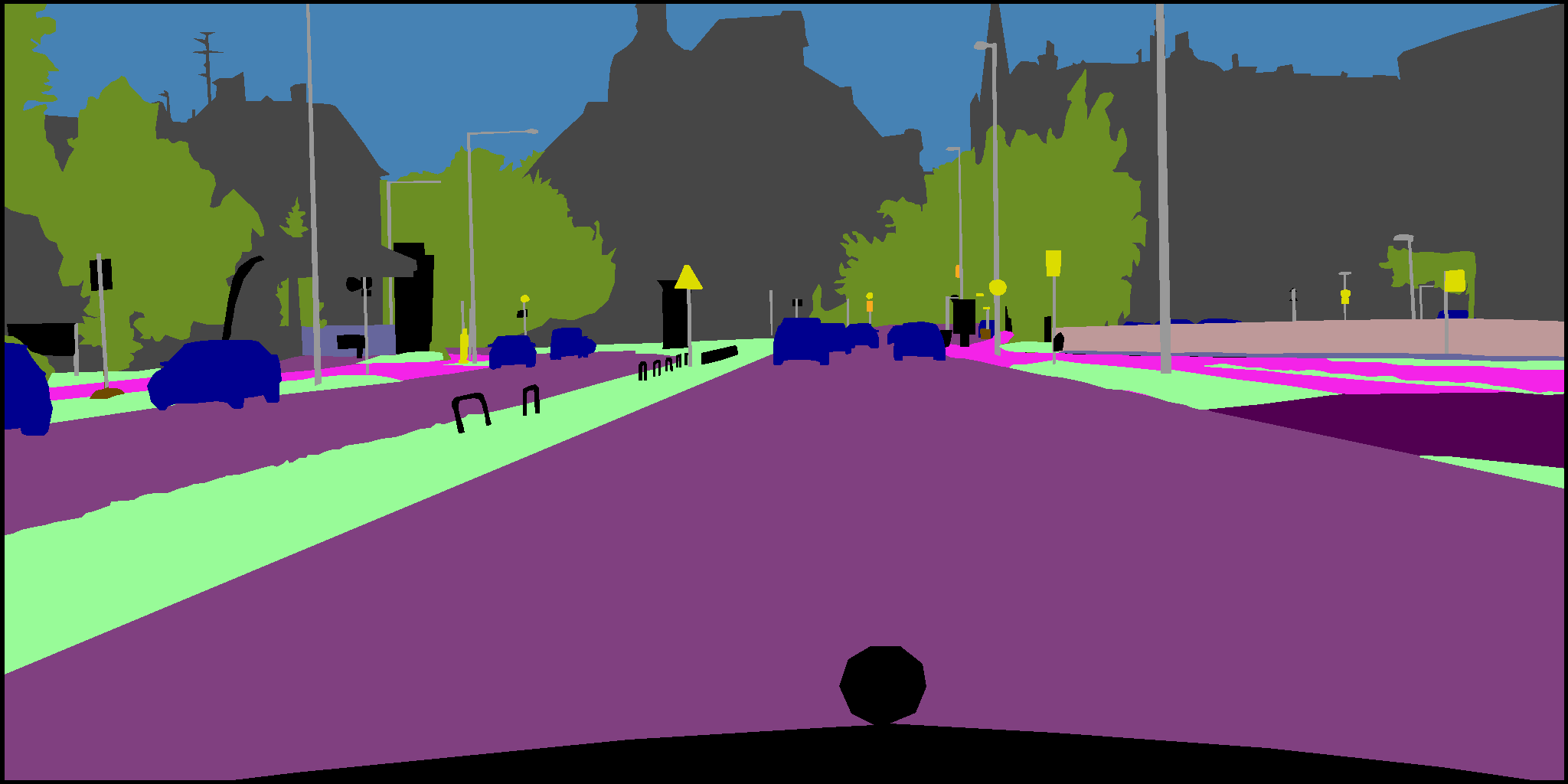}}
\end{minipage}
}

\caption{Visual image segmentation results for cityscape val dataset. The six columns from left to right are the input image, the boundary loss results from SANet-S, the heat map results from SANet-S, the auxiliary loss results from SANet-S, the prediction results from SANet-S, and the ground truth.}
\label{output_image}
\end{figure*}

\begin{table}[H]
\caption{Ablation experiment of APPPM. Except for the distinction in the pyramid pooling module, the remaining components utilize Baseline + APPPM + SAD from the network architecture's ablation experiments as a foundation.}
\begin{center}
\renewcommand\arraystretch{1.3}
\label{apppm}
\begin{tabular}{|l|c|c|c|c|}
\hline
Module & Speed(FPS) & Params(M) & mIOU(\%)       & Year \\ \hline
PPM    & 72.2       & 6.16      & 75.3           & 2017 \\ \hline
DAPPM  & 58.6       & 6.84      & 77.7           & 2022 \\ \hline
SPPM   & 67.2       & 6.11      & 77.4           & 2022 \\ \hline
AAPPM  & 62.8       & 6.64      & 77.8           & 2023 \\ \hline
APPPM  & 65.1       & 8.26      & \textbf{78.1} & -   \\ \hline
\end{tabular}
\end{center}
\end{table}

\paragraph {\textbf{APPPM}}
The pyramid pooling module has the ability to extend the model's sensory field and perceive global and local information. We compared APPPM with other modules with similar functionality, such as PPM. The results are shown in Table~\ref{apppm}, where the mIoU of APPPM is improved by 3$\%$ compared to PPM. We also have a small improvement over other recently proposed pyramid pooling module. In summary, APPPM improves the performance of the network from 75.3$\%$  to 78.1$\%$  and performs the best in comparison with other semantic extraction modules.

\paragraph {\textbf{Loss Function}}
The choice of an appropriate loss function plays a crucial role in deep learning tasks. We compare two loss functions commonly used for semantic segmentation: cross-entropy loss and online hard example mining cross-entropy loss. Meanwhile, to improve the model's ability to perceive spatial information, we also add BoundaryLoss at different locations of the model. The results show that the online hard instance mining cross-entropy loss function has better performance compared to the standard cross-entropy loss function, and the accuracy of the model is further improved after adding BoundaryLoss. The results of the comparison are shown in Table~\ref{loss}. Figure~\ref{output_image} shows the output image of the auxiliary segmentation header used for BoundaryLoss.

\begin{table}[]
\caption{Ablation study of losses}
\begin{center}
\renewcommand\arraystretch{1.5}
\label{loss}
\begin{tabular}{ccccc}
\hline
\multicolumn{2}{c}{Main Loss} & 
\multicolumn{2}{c}{Extra Loss} &
\multirow{2}{*}{mIOU(val)}\\
\cmidrule(r){1-2}\cmidrule(r){3-4}
CE          & OHEM CE         & Auxiliary Loss & Broundary Loss &                     \\\hline
\checkmark  &                 &                &                & 76.2                  \\\hline
            & \checkmark      &                &                & 76.8                  \\\hline
            & \checkmark      & \checkmark     &                & 77.4                  \\\hline
\checkmark  &                 & \checkmark     & \checkmark     & 78.2                  \\\hline
            & \checkmark      & \checkmark     & \checkmark     & 78.6      \\\hline           
\end{tabular}
\end{center}
\end{table}

\subsection{\textbf{Comparison}}

\paragraph {\textbf{CityScapes}}
In Table~\ref{table4}, we present a comparison of SANet and previous state-of-the-art real-time semantic segmentation models using the Cityscapes dataset. TensorRT, an open-source library optimized for high-performance deep learning inference, is utilized. With the acceleration provided by TensorRT, network models for real-time semantic segmentation can typically achieve more than a twofold increase in speed. However, it should be noted that many research papers do not specify whether or not they utilize TensorRT in model speed comparisons. To ensure the fairness of speed comparison between different models, we follow the speed comparison guidelines proposed by Li et al~\cite{li2022holoparser}. and list the speeds of both Pytorch and TensorRT methods separately.

In terms of comparing model accuracy, the evaluation of the Cityscapes test dataset typically involves two methods. One way is to train the model on the training and validation datasets and then generate the results of the test dataset and submit them to the server for evaluation. The other way is to train using the train dataset and implement a multi-scale approach to generate the test dataset results, which are then submitted to the server for evaluation. We utilized the multi-scale comparison specification presented by Leonel et al~\cite{rosas2021fassd}., which comprises four scales (0.5, 0.75, 1, 1.25).DDRNet and PIDNet are the current state-of-the-art models in the realm of real-time semantic segmentation. For our specific comparison, we employed the aforementioned multi-scale comparison specification to evaluate the performance of DDRNet and PIDNet exclusively. It should be noted that the comparison is conducted by utilizing the officially provided model weights, producing predicted images of the test dataset using both traditional and multi-scale methods, and submitting these outcomes to the evaluation server for assessment.

\begin{table}[]
\caption{Comparison with state-of-the-art real-time methods on the CamVid test dataset.}
\begin{center}
\renewcommand\arraystretch{1.3}
\label{camvid}
\begin{tabular}{lccc}
\hline
Model       & Speed(FPS) & mIOU(\%)       & Year \\ \hline
MSFNet\cite{si2019real}      & 91         & 75.4           & 2019 \\
BiSeNetV2\cite{yu2021bisenet}   & 124        & 72.4           & 2020 \\ 
BiSeNetV2-L\cite{yu2021bisenet} & 32.7       & 73.2           & 2020 \\ 
STDC2-Seg\cite{fan2021rethinking}   & 152.2      & 73.9           & 2021 \\ 
HyperSeg-S\cite{nirkin2021hyperseg}  & 38         & 78.4           & 2021 \\ 
DMANet\cite{weng2022deep}      & 119.8      & 76.2           & 2022 \\ 
DDRNet\cite{pan2022deep}      & 230        & 78.6           & 2022 \\ 
DMRNet\cite{wang2023deep}      & 96.1       & 75.8           & 2023 \\ \hline
SANet-S     & 147        & \textbf{78.8} & -    \\ \hline
\end{tabular}

\end{center}
\end{table}

\paragraph {\textbf{CamVid}}
To further validate the SANet model's performance, we compared it with other models utilizing the Camvid dataset. Similar to previous studies, we employed the SANet model's pre-training weights from the Cityscapes dataset. Furthermore, the model's training and inference input resolution was established at 960×720. As depicted in Table~\ref{camvid}, SANet attained 78.8$\%$ mIOU accuracy on the CamVid test dataset. In contrast to other models, SANet delivers the highest accuracy. Overall, SANet achieves strong performance on the Camvid dataset, effectively balancing the trade-off between precision and speed.
\section{Conclusion}
In this paper, we try to merge the encoder-decoder structure with the two-pathway approach and thus propose the SANet. While there is typically significant attention paid to size and shape variations of the convolutional layer, less thought is given to variations of the pooling layer. To improve the extraction of semantic information, we introduce an asymmetric pooling layer and design an asymmetric pooling pyramid pooling module. Decoders serve to recover features in semantic segmentation. Simple Attention Decoder helps SANet utilize both horizontal and vertical attention for feature fusion and recovery while maintaining inference speed. Experimental results from two urban streetscape benchmark datasets, namely Cityscape and CamVid, demonstrate the efficiency and effectiveness of our proposed SANet.
In future studies, we plan to enhance the model's speed and accuracy by optimizing the number of channels and configuring the convolutional blocks properly. Moreover, we aim to employ SANet in diverse scenarios, including medical image processing and remote sensing image analysis.


\bibliographystyle{ieeetr}
\bibliography{sanet.bib}

\end{document}